\pdfoutput=1
\documentclass[11pt]{article}
\usepackage[final]{acl}
\usepackage{times}
\usepackage{latexsym}

\usepackage[T1]{fontenc}
\usepackage[utf8]{inputenc}

\usepackage{microtype}

\usepackage{inconsolata}

\usepackage{graphicx}
\usepackage{CJKutf8}
\usepackage{times}
\usepackage{soul}
\usepackage{url}
\usepackage{amsmath}
\usepackage{amsthm}
\usepackage{booktabs}
\usepackage{algorithm}
\usepackage{algorithmic}
\usepackage[switch]{lineno}
\usepackage{booktabs}
\usepackage{tabularx}
\usepackage{amsfonts}
\usepackage{makecell}
\usepackage{amssymb}
\usepackage{multirow}
\usepackage{microtype}
\usepackage{soul}
\usepackage{booktabs}
\usepackage{enumitem}
\usepackage{pifont}
\usepackage{colortbl}
\usepackage{tikz}
\usepackage{soul}
\usepackage{graphicx}
\usepackage{microtype}
\usepackage{array}
\usepackage{bbding}
\usepackage{xspace,mfirstuc,tabulary}
\definecolor{c1}{HTML}{DB7093}
\definecolor{c2}{HTML}{C71585}
\definecolor{c3}{HTML}{006400}

\newif\ifdraft
\drafttrue

\title{CausalAbstain: Enhancing Multilingual LLMs with Causal Reasoning for Trustworthy Abstention}

\author{
Yuxi Sun$^{1}$\thanks{Equal contribution.}, 
Aoqi Zuo$^{2}$\footnotemark[1],
\textbf{Wei Gao$^{3}$},
\textbf{Jing Ma$^{1}$}\thanks{Corresponding author.}\\
$^{1}$ Department of Computer Science, Hong Kong Baptist University \\
$^{2}$ School of Mathematics and Statistics, The University of Melbourne \\
$^{3}$ School of Computing and Information Systems, Singapore Management University \\
\href{mailto:csyxsun@comp.hkbu.edu.hk}{csyxsun@comp.hkbu.edu.hk}, ~\href{mailto:azuo @student.unimelb.edu.au}{azuo@student.unimelb.edu.au}\\ ~\href{mailto:weigao@smu.edu.sg}{weigao@smu.edu.sg}, ~\href{mailto:majing@comp.hkbu.edu.hk}{majing@comp.hkbu.edu.hk}
}

\begin{document}
\begin{CJK*}{UTF8}{gbsn}
\maketitle

\newcommand{\mytextbf}[1]{{%
  % \smaller 命令减小字体大小，你可以根据需要调整
  % \smaller
  \fontsize{10.5pt}{<baselineskip>}\selectfont
  % 如果需要调整字母间距，可以使用 \setstretch 命令（来自 setspace 包）
  % \setstretch{1.05} % 例如，增加间距（这里只是示例，通常不需要）
  % 但由于我们要减小间距，可以通过其他方式实现，比如使用 \kern
  % 不过，直接调整字母间距通常不是好的做法，因为它会影响可读性
  % 这里我们仅减小字体大小
  \textbf{#1}%
}}
% \begingroup
% \setlength{\spaceskip}{-0.5em} % 设置单词之间的空格距离
% \setlength{\xspaceskip}{0.5em} % 设置标点符号后的空格距离
% \textbf{This is a test sentence with reduced space between words.}
% \endgroup

% Ours
\maketitle
\begin{abstract}
Large Language Models (LLMs) often exhibit knowledge disparities across languages. Encouraging LLMs to \textit{abstain} when faced with knowledge gaps is a promising strategy to reduce hallucinations in multilingual settings. Current abstention strategies for multilingual scenarios primarily rely on generating feedback in various languages using LLMs and performing self-reflection. However, these methods can be adversely impacted by inaccuracies and biases in the generated feedback. To address this, from a causal perspective, we introduce \textit{CausalAbstain}, a method that helps LLMs determine whether to utilize multiple generated feedback responses and how to identify the most useful ones. Extensive experiments demonstrate that \textit{CausalAbstain} effectively selects helpful feedback and enhances abstention decisions with interpretability in both native language (\textsc{Casual-native}) and multilingual (\textsc{Causal-multi}) settings, outperforming strong baselines on two benchmark datasets covering encyclopedic and commonsense knowledge QA tasks. Our code and data are open-sourced at \url{https://github.com/peachch/CausalAbstain}.
%...
\end{abstract}
\section{Introduction}
Large language models (LLMs) demonstrate impressive capabilities in encoding vast amounts of information and supporting knowledge-intensive tasks~\cite{petroni-etal-2021-kilt, yu2023kola, yang2024survey, yang2025multimodal, shang2024survey}. However, hallucinations and bias~\cite{mishra2024finegrainedhallucinationdetectionediting, kumar-etal-2023-language,Ji_2023} can arise when knowledge is missing or inaccurate, posing challenges to the reliability of LLMs~\cite{feng-etal-2024-teaching}. A promising approach, which has inspired recent research, is to teach LLMs to \textit{abstain} --- \textit{avoid incorrect answers in low-confidence scenarios to mitigate hallucinations and factual inaccuracies}~\citep{madhusudhan2024llmsknowanswerinvestigating, feng-etal-2024-dont,feng-etal-2024-dont}. 
\begin{figure}[t!]
    \begin{center}
    % \centerline{\includegraphics[width=1\columnwidth, height=8cm]
    \centering{\includegraphics[width=0.9\columnwidth]{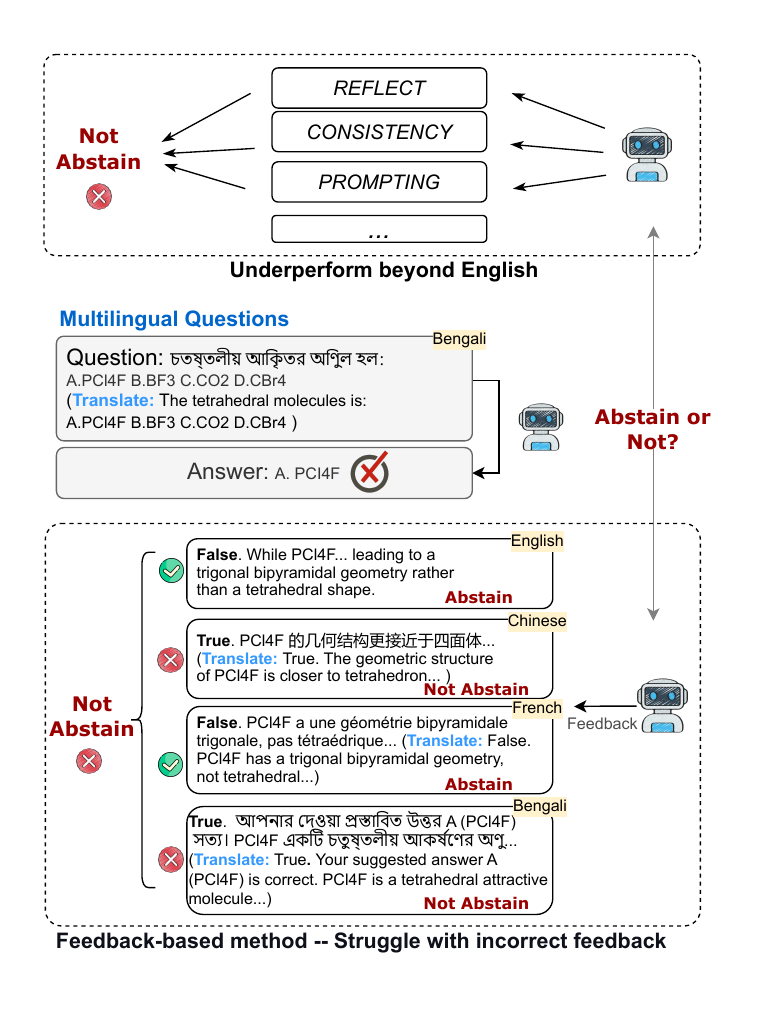}}
    \vspace{-1em}
    \caption{Prior approaches often underperform beyond English, particularly in low-resource languages. While the feedback-based method is well-suited for multilingual queries, some generated feedback may be incorrect (e.g., Chinese and Bengali), leading to incorrect decisions, i.e., failing to abstain from a wrong answer (A). The English translations (labeled as \textcolor{cyan}{Translate:}) are provided for reading.}
    \vspace{-1em}
    \label{fig: example}
    \end{center}
\end{figure}

Existing studies have explored abstention strategies in English~\cite{varshney2023postabstentionreliablyreattemptingabstained,feng-etal-2024-dont,yang2024alignmenthonesty}, but the ability of LLMs to abstain in multilingual settings remains largely underexplored~\cite{feng-etal-2024-teaching}. The factual accuracy of multilingual LLMs, particularly in low-resource languages, is often poorer~\cite{zhang-etal-2023-dont, kang2024comparinghallucinationdetectionmetrics}, limiting their global applicability and leaving many regions underserved. Therefore, developing robust abstention strategies across languages is essential to enhancing the reliability and inclusivity of LLMs.

Current \textit{training-based}~\cite{azaria-mitchell-2023-internal,slobodkin-etal-2023-curious} and \textit{calibration-based}~\cite{zhou2024batchcalibrationrethinkingcalibration,liu2024litcablightweightlanguagemodel,tian-etal-2023-just} strategies require a hold-out dataset for hyperparameter tuning and determining confidence score thresholds, which may hinder generalization across different knowledge domains.
%and \textit{calibration-based} strategies utilize a confidence score threshold for abstain, both struggles in low-resource languages~\cite{feng-etal-2024-teaching}. 
\textit{Prompt-based} abstention strategies~\cite{kadavath2022languagemodelsmostlyknow,edunov2018understandingbacktranslationscale}, such as prompting LLMs to evaluate their own answers before abstaining or generating additional knowledge prior to answering, have been studied. Similarly, \textit{self-consistency} method~\cite{wang2023selfconsistencyimproveschainthought}
%\textit{collaboration-based}~\cite{feng-etal-2024-dont, wang2023selfconsistencyimproveschainthought} abstain approaches 
encourages LLMs to reflect on their responses cooperatively. These approaches have been shown to underperform in low-resource languages, demonstrated by~\citet{feng-etal-2024-teaching}. To address this limitation, they propose a \textit{feedback-based method} to teach LLMs to abstain by generating and reflecting on relevant feedback, making it more adaptable to multiple languages.

% \vspace{-0.3em}
However, all of these methods still raise concerns because: 1) They rely solely on LLMs evaluating their own generated texts, which can be affected by hallucinations and potential bias~\cite{xie2024adaptivechameleonstubbornsloth, Ji_2023}; 2) The generated information or feedback may be irrelevant or unreliable, particularly in different language resources, leading to negative impacts. Figure~\ref{fig: example} illustrates this issue: When LLM reviews its previous answer and provides feedback in different languages, including native language\footnote{The same language as the original answer is referred to as ``native language".}, some feedback is of low quality or even incorrect in certain languages, ultimately influencing the final abstention decision (e.g., the Chinese feedback in the figure). This raises a crucial research question: \textit{How can we determine whether to utilize LLM-generated feedback and how to identify the helpful ones from multiple feedback sources?} 

Inspired by the ability of the Structural Causal Model (SCM)~\cite{pearl2000models} to measure path-specific causal effects, we propose a causal approach to assess the impact of feedback on a model's proposed answer. This allows us to determine whether to utilize the generated feedback for answer refinement and to identify the most helpful feedback to make a final abstention decision.
%from multi-lingual languages. %Causal ... can ...., so we expected.... 
Specifically, we probe a multilingual LLM to provide feedback on its proposed answer in both its native language (\textsc{Casual-native}) and multiple related languages\footnote{Following the definition of language relatedness by~\citet{sun-etal-2021-cross}, which considers cultural, geographical, and typological factors.} (\textsc{Casual-multi}). By generating feedback over multiple runs, the model can produce diverse knowledge and varying assessments of the answer. We then apply causal inference to evaluate the causal effect of the generated feedback, determining whether it meaningfully enhances the abstention decision. Additionally, since different languages have varying degrees of representation in LLM pre-training data --- some being severely underrepresented~\cite{lai2023okapiinstructiontunedlargelanguage}, we propose candidate aggregation to come up with a more robust multilingual abstention strategy, which employs a voting mechanism to ensure a reliable abstention decision across multiple languages.

Our contributions are mainly three-fold:
\begin{itemize}[leftmargin=*]
\item We present the first study on causal abstention in multilingual LLMs and propose \textit{CausalAbstain}, a training-free approach that helps multilingual LLMs identify incorrect or biased feedback and abstain accordingly.  
\item Our method integrates feedback from both the native language and multiple related languages (i.e., \textsc{Causal-native} and \textsc{Causal-multi}), leveraging path-specific causal effect measurement to guide feedback selection. 
\item We evaluate \textsc{Causal-multi} and \textsc{Causal-native} on two datasets covering encyclopedic and commonsense knowledge QA. Experimental results demonstrate that our approach outperforms state-of-the-art baselines in abstention effectiveness.
\end{itemize}

% \section{Methodology}
\section{Causality with Feedback}
In this section, we introduce the fundamental concepts of causal inference, providing the background that supports \textit{CausalAbstain} in \S\ref{section:CausalAbstain}.

\subsection{Causal Graph}
\begin{figure}[t!]
    \begin{center}
    % \centerline{\includegraphics[width=0.8\columnwidth]{imgs/graph (7).pdf}}
    \centerline{\includegraphics[width=0.7\columnwidth]{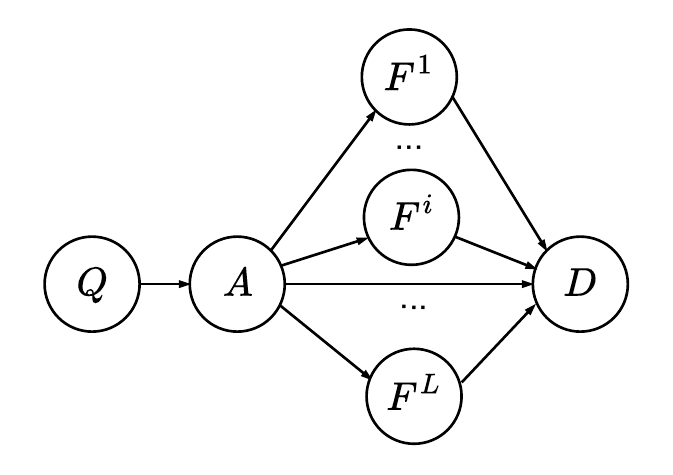}}
    \caption{The causal graph of the question-answering problem with feedback.}
    %$Q$ and $A$ represent the question and the original answer, respectively. $F^i$ denotes the feedback provided in languages $i$. Suppose the number of language for feedback is $L$ and let $D$ represents the final decision.
    \vspace{-2em}
    % \label{fig: example}
    \label{fig: causal graph}
    \end{center}
\end{figure}

Causal inference is typically performed using the Structural Causal Model~\cite{pearl2000models}. A key component of SCM is the causal graph, a directed acyclic graph (DAG) that represents causal relationships between variables. We denote the causal graph as $\mathcal{G}=\{\mathcal{V},\mathcal{E}\}$, where $\mathcal{V}$ represents the set of variables in the graph and $\mathcal{E}$ denotes the set of causal relationships between them. To address the question answering with feedback problem, we construct a specific causal graph in Figure~\ref{fig: causal graph}. 

In the abstention task, the question ($Q$) and the originally proposed answer ($A$) together form the input to the LLM, which then generates the feedback $F^i$ in language $i$. In Figure~\ref{fig: causal graph}, a directed edge from one variable to another indicates a causal relationship, meaning the former influences the latter. The \textit{likelihood distribution of the final abstention decision}, denoted as $D$, is influenced by both $A$ and the generated feedback $F^i$. In the absence of feedback, $D$ is directly decided by $A$. Additionally, since multilingual feedback can affect $D$, multiple causal paths may exist from $A$ to $D$, mediated by different feedback variables in $\{F^i|1 \le i \le L\}$, where $L$ represents the number of languages considered for feedback\footnote{We do not model a direct causal link from $Q$ to $F^i$, as the question alone cannot trigger feedback on the answer— abstention feedback is only meaningful in the context of the answer. The same reasoning applies to the absence of a direct edge from $Q$ to $D$.}.

\subsection{Causal Effect}\label{section:CausalEffect}
Based on the SCM, various causal effects can be measured to quantify the impact of a treatment on an outcome. In Figure~\ref{fig: causal graph}, $A$ serves as the treatment for $D$. There are two distinct types of effects between $A$ and the final decision $D$: the natural direct effect (NDE) (e.g., $A \rightarrow D$) and the indirect effect mediated by the generated feedback (e.g., $A \rightarrow F^i \rightarrow D$). The total effect (TE) of $A$ on $D$ can be decomposed into NDE and total indirect effect (TIE), as follows: 
\begin{equation}
    \mathrm{TE}= \mathrm{NDE} + \mathrm{TIE}.
\end{equation} 
If we denote the TIE with respect to each feedback $F^i$ as $\mathrm{TIE}^i$, then 
\begin{equation}
    \mathrm{TIE} = \sum_{i=1}^L \mathrm{TIE}^i.
\end{equation} 
NDE is computed by fixing the feedback mediations and comparing the potential outcomes with and without $A$:
\begin{equation} \label{eq:NDE}
\mathrm{NDE}=\mathbb{E}[D(Q, A)-D_0],
\end{equation}
where $\mathbb{E}$ represents the expectation operator, and $D(Q, A)$ represents the final decision based solely on the original answer given the question,
%\yx{if $A$ means question together with original answer, there seems not consistency?}
meaning it corresponds exactly to the original answer. $D_0$ serves as the baseline for $D$, which is typically modeled as a binomial distribution with a binominal probability 0.5.
The indirect effect mediated by feedback $F^i$ is computed by contrasting the potential outcomes with and without the feedback $F^i$, given the presence of $A$:
\begin{equation} \label{eq:TIE^l}
\mathrm{TIE}^i=\mathbb{E}[D(Q, A, F^i)-D(Q, A)].
\end{equation}

%%%%%%%%%%%%%%%%%%%%%%%%%%%%%%%%%%%%%%%%%%%%%%%%%%%%%%%%%%%%%%%%%%%%%%%%%%%%%%%
\section{CausalAbstain}\label{section:CausalAbstain}

\begin{figure*}[tbh!]
    \begin{center}
    \centerline{\includegraphics[width=1.025\textwidth]{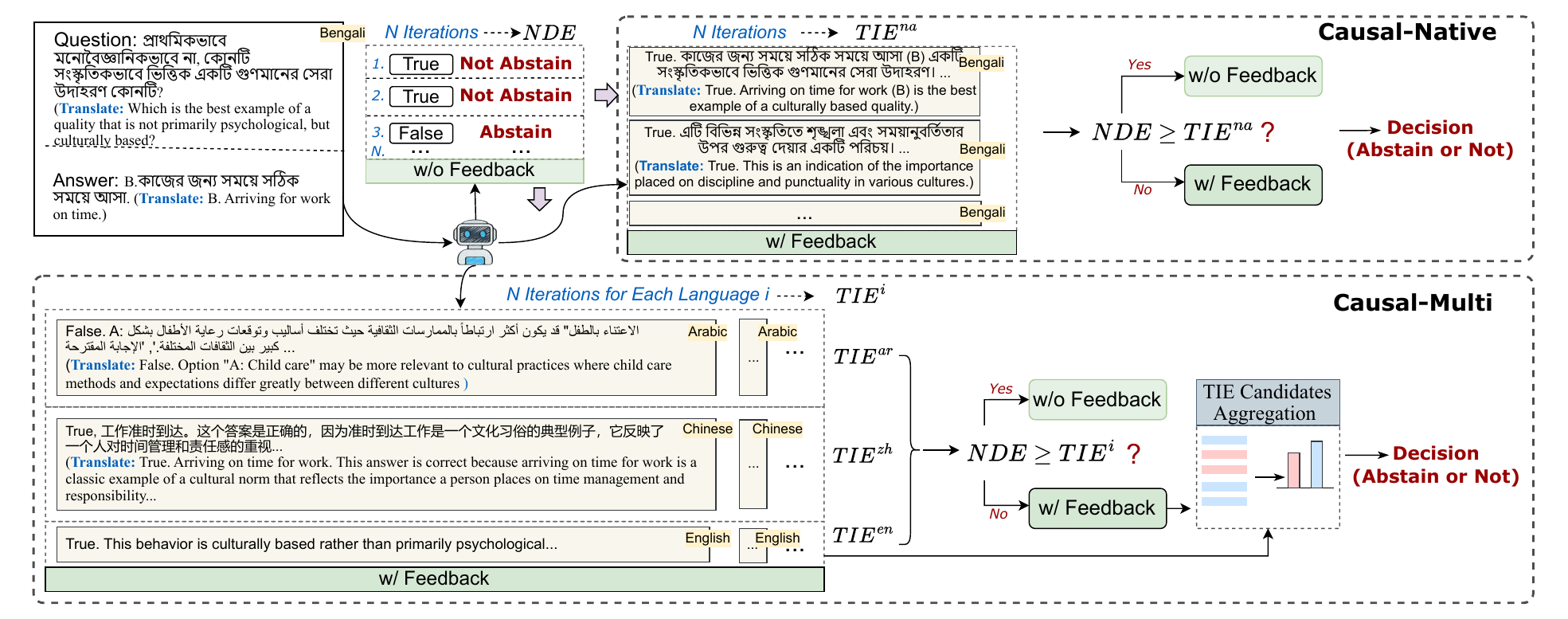}}
    \caption{The framework of \textit{CausalAbstain}. We propose an abstention strategy to determine whether to adhere to or abstain from the proposed answer (i.e., the answer B) based on the feedback in the native language (\textsc{Causal-native}) and related languages (\textsc{Causal-multi}), by leveraging natural direct effect (NDE) and total indirect effect (TIE) to assess the impact of feedback on the final abstention decision.  We provide the English translations (\textcolor{cyan}{Translate:}) for reading.}
    \vspace{-2em}
    \label{fig: framework}
    \end{center}
\end{figure*}

Ideally, feedback can provide valuable insights to improve an answer to a question. However, not all feedback is helpful. As illustrated in Figure~\ref{fig: example}, feedback may sometimes be less informative, introduce erroneous information, or even be incorrect, which can negatively impact the abstention performance. 
To address this challenge, we propose a causality-based abstention mechanism to determine whether to incorporate feedback and how to effectively utilize it in deriving the final decision.

\subsection{Task Formulation}
% \aq{how to formulate this section?}
We focus on assisting multilingual LLMs in abstaining during question answering~\cite{feng-etal-2024-dont}. The LLM is first prompted with a given query $Q$, which may be in different languages, and provides the answer $A=\operatorname{LLM}(Q)$. The LLM is then tasked with providing an abstention decision based on the proposed answer and query: $f(Q, A)\rightarrow \{\textit{Abstain}, \textit{Not Abstain}\}$.
%\wei{Would it be more straightforward to output abstain or not abstain instead of True or False?}. 
For our setting, the prompt to elicit LLMs' abstention decision directly is: ``Please review the correctness of proposed answer \textit{True} or \textit{False} directly."
The LLM is expected to \textit{Abstain} when it is likely to provide an incorrect answer and \textit{Not Abstain} when it is expected to answer correctly.

When incorporating multilingual feedback $F^i$, where $F^i = \operatorname{LLM}(Q, A|i)$ represents the feedback generated in language $i$, the feedback is elicited using the prompt~\cite{feng-etal-2024-teaching}: ``Please review the proposed answer and provide a paragraph of feedback on its correctness. Feedback should be in language $i$". The abstention decision is then made based on the multilingual feedback, with the decision function updated to $f(Q, A,{F^1, F^2... F^L}) \rightarrow \{\textit{Abstain}, \textit{Not Abstain}\}$.
% \wei{The same question}. 
Specifically, for the feedback in each language, the prompt is: ``Based on the feedback for measuring the correctness of the answer, is the proposed answer \textit{True} or \textit{False}?"

Our goal is to develop a robust abstention strategy $f$ by leveraging causality. The framework of \textit{CausalAbstain} is illustrated in Figure~\ref{fig: framework}, which consists of methods for incorporating native-language (i.e., \textsc{Causual-native}) and multilingual (i.e., \textsc{Causual-multi}) feedback, presented in \S\ref{section:native} and \S\ref{section:multi}, respectively.
\subsection{\textsc{Causual-native}}\label{section:native}
For the native language, we propose \textsc{Causal-native} to determine whether to adhere to the proposed answer or abstain from it based on the monolingual feedback. %by leveraging the SCM for its fine-grained measurement capabilities.
The causal graph for our problem setting follows Figure~\ref{fig: causal graph}, where we consider two specific types of causal relationships between the question $Q$,  the original answer $A$, and the final decision $D$: 1) $\mathrm{NDE}$ implied by $A \rightarrow D$. It quantifies the direct impact of $A$ on $D$ in the absence of feedback. 
2) $\mathrm{TIE}^{na}$ implied by causal path $A \rightarrow F^{na} \rightarrow D$, which is mediated by feedback $F^{na}$ in the native language. It captures the extent to which feedback contributes to the overall causal effect of $A$ on the final decision $D$.

Given the general Equations~\eqref{eq:NDE} and~\eqref{eq:TIE^l}, we use the Jensen-Shannon Divergence (JSD) in practice to compare the different outcome distributions in these equations. For the expectation operation, given each query-answer pair, we iteratively generate $N$ feedback instances in the native language to approximate $F^{na}$, denoted as $\hat{F}^{na}$. We then use it to compute $\hat{D}(Q, A, \hat{F}^{na})$. Additionally, we prompt without feedback in $N$ iterations to approximate $\hat{D}(Q, A)$. 
For the $j$-th feedback instance, we denote the corresponding decision as $\hat{D}_j(Q,A, \hat{F}_j^{na})$. 
All feedback are thus collected as $\hat{D}(Q, A,\hat{F}^{na})=\{\hat{D}_j({Q}, {A},\hat{F}_j^{na})|1\le j \le N\}$. Therefore, the practical computation of $\mathrm{NDE}$ and $\mathrm{TIE}^{na}$ is formulated as follows:
%\wei{just wonder why there is no expectation operator below, as eq 3,4.}
\begin{equation}
  \mathrm{NDE}=\operatorname{JSD}(\hat{D}(Q, A),D_0), 
\end{equation} 
\begin{equation}
    \mathrm{TIE}^{na}=\operatorname{JSD}(\hat{D}(Q, A,\hat{F}^{na}),\hat{D}(Q, A)).
\end{equation}

For any given example, our causal abstention strategy $f$ determines whether to incorporate feedback based on the relationship between $\mathrm{NDE}$ and $\mathrm{TIE}^{na}$, as illustrated in Figure~\ref{fig: framework}. If $\mathrm{NDE} \geq \mathrm{TIE}^{na}$, 
the indirect effect is minimal, indicating that the feedback has little to no impact on improving the response. In this case, the abstention decision is made without considering the feedback and thus majority voted by $\hat{D}(Q, A)$. 
Conversely, if $\mathrm{NDE} < \mathrm{TIE}^{na}$, the indirect effect is substantial, suggesting that the feedback plays a significant role in refining the response. In this case, the abstention decision is made considering the feedback and is voted by majority of $\hat{D}(Q, A, \hat{F}^{na})$ over $N$ native feedback instances. 

\subsection{\textsc{Causal-multi}}\label{section:multi}
Previous research has demonstrated that multiple feedback items in related languages can help LLMs identify knowledge gaps across diverse languages, cultures, and communities~\cite{feng-etal-2024-teaching}. However, the challenge remains in effectively determining whether to utilize multilingual feedback. To address this, we formulate the abstention strategy \textsc{Causal-multi} for multilingual settings.

The calculation of $\mathrm{NDE}$ and $\mathrm{TIE}^i$ for feedback in language $i$ follows the same approach as described in \S\ref{section:native}. And the abstention strategy $f$ is: If $\mathrm{NDE} \geq \mathrm{TIE}^i$  for feedback in any languages $i$, it implies that the feedback can be disregarded, and the original answer can be used as the final answer --- similar to the native language scenario. Conversely, if there exists any case where $\mathrm{NDE} < \mathrm{TIE}^i$ in some language $i$, it suggests that the feedback could play a crucial role, such as identifying key nuances or providing knowledge that significantly enhances the response. 

Considering the inconsistent representation of different languages, especially low-resource ones in the pre-training data of LLMs~\cite{lai2023okapiinstructiontunedlargelanguage}, we propose a robust multilingual abstention strategy to mitigate potential biases introduced by feedback in a specific language. To achieve this, we
introduce an aggregated voting mechanism across feedback from all languages. Specifically, we consider $L$ languages and denote the abstention decision provided by the feedback in language $i$ as $\mathbf{\hat{D}}^i$, where $\mathbf{\hat{D}}^i=\{\hat{D}^i_1, \hat{D}^i_2,\ldots, \hat{D}^i_N\}$ and $N$ is the number of feedback instances. The final decision is derived from the set $\{\mathbf{\hat{D}}^1, \mathbf{\hat{D}}^2, \ldots, \mathbf{\hat{D}}^L\}$, with a majority voting crossing over the $N \times L$ choices to determine whether to abstain or not. 

\setlength{\tabcolsep}{1.4pt} % Default value: 6pt
\renewcommand{\arraystretch}{1.02} % Default value: 1
\begin{table*}[ht!]
\center
\small
\begin{tabular}{lcccccccc>{\columncolor{gray!15}}c|cccccccc>{\columncolor{gray!15}}c} \\\toprule
\multirow{2}{*}{\textbf{Method}} & \multicolumn{9}{c}{\textbf{M-MMLU}} & \multicolumn{9}{c}{\textbf{M-Hellaswag}} \\\cline{2-19}
& zh& it & ar& id & bn & te & ne & kn & $\mathrm{Overall}$ & zh& it &ar& id & bn & te & ne & kn & $\mathrm{Overall}$ \\ \hline
&\multicolumn{15}{c}{\textbf{\textsc{Aya-13B}}} \\ \hline
\textsc{Ask Cali.}&\underline{.552}& .534 &\textbf{.593}   &\textbf{.547}  & \textbf{.587} &.424 &.477 & .539  &.532&  .511&.504  &.520  &.541 & .481 & .538 &.498& .496 & .511\\ 
% \textsc{Instruct} &  &&  &   & &  &  &  & & &  &  & &  &  &  &  &   \\ 
\textsc{Reflect}   &.478& .474 &.469   & .463 & .370 &.395  &.320  &.418  & .423&   .481 &.483 &.507  & .481  &.463  &.502 &.472&.509 & .477\\ 
\textsc{Moreinfo}  &.500& .479 &.500   & .435& .348 &.429  &.353  &.411  & .431&  .489 &.457 & .533 & .519 &.506  &.552 & .429&.452& .489 \\ 
\textsc{Conflict}  & .470  & \textbf{.547}  & .456  &.461 & .550 & .497 &.550  &.500  & .504& .485 &\underline{.538} &.515  &.511  &.472  & .484& .498&.496 & .501 \\ 
\textsc{Multi-Related}& \textbf{.555} & .525  &  \underline{.560} & .510 & .555 & \underline{.524} & \textbf{.555} & \textbf{.609}  & \textbf{.549}&  .516  &.517 &.511 &.455  &.580  &\underline{.557} & .494&.592 &.525  \\ \hline
\textsc{Causal-native}  &.482  & \underline{.538}  & .510  &.475 &.550  &.497  &.475  &.507  & .504 & \textbf{.593}&.534  &\underline{.580}  & \underline{.613} &\textbf{.679} &.457 &\underline{.552}& \underline{.657}  & \underline{.564} \\ 
\textsc{Causal-multi} &.512&\underline{.538}  & .530  & \underline{.535} & \underline{.574} & \textbf{.544}  & \underline{.545} & \underline{.579}  &\underline{.545}  & \underline{.588} &\textbf{.564}  &\textbf{.589}  & \textbf{.626} & \underline{.670}& \textbf{.574}& \textbf{.597} & \textbf{.685} & \textbf{.612}\\ \hline
\textbf{}&\multicolumn{15}{c}{\textbf{\textsc{GPT-4o}}} \\ \hline
\textsc{Ask Cali.} &.208  &.205& .293 & .208  & .462& .413 &.419  &.531   & .343&.377& .192 & .336 &.265 & .351 &.524  & .395 &.487 &.366  \\ 
% \textsc{Instruct} &  &&  &   & &  &  &  & & &  &  & &  &  &  & &   \\ 
\textsc{Reflect} & .503 &.526&.503  &.552   &.462 & .465 & .493 &  .527 & .504& .279 &.256&.368&.381&.383&.513& .413 &.526 &.390 \\
\textsc{Moreinfo} & .768 &\underline{.833} &.682  & .779 &.613 &.597  & \underline{.669} & .543  &.686 &.617  & .660 & .704&.671  & .589 & .551 &.516 & .441  &.594 \\ 
\textsc{Conflict} & .665 &.692& .623 &  .734 &  .503 &.519  &.487 & .565  & .599& .416 &.417  &.507 & .471 &.610  & .585 &.561 & .599  &.521\\ 
\textsc{Multi-Related}& \underline{.770} & \textbf{.850}& \underline{.740} &\underline{.800}  &.720 &.605  &\textbf{.675}  &\underline{.639}   &\underline{.725} & .766   & .769&\textbf{.776}  &.761 &\underline{.747} &.626 &.703 &\textbf{.691}  & .730  \\ \hline
\textsc{Causal-native} & \textbf{.805} & .756  & \textbf{.760}  & \textbf{.805}& \underline{.737} &\underline{.610}  & .608 &.623 & .713  & \textbf{.792} & \underline{.833} &.730 & \underline{.800} &.746  &\textbf{.721} &\underline{.741} &.631 &\underline{.749}  \\
\textsc{Causal-multi}& .765 &.830 &  \textbf{.760} & .795 & \textbf{.740}&\textbf{.660}  &\textbf{.675}  &\textbf{.671} &\textbf{.738}  &  \underline{.772}&\textbf{.846}  &\underline{.743}  & \textbf{.806} & \textbf{.753}& \underline{.700}&\textbf{.748} &\underline{.664} & \textbf{.754} \\ \hline
\textbf{}&\multicolumn{15}{c}{\textbf{\textsc{ChatGPT}}} \\ \hline
\textsc{Ask Cali.}  &.440& .385 & \textbf{.535}  &.427 & \textbf{.618} & \textbf{.622} &\underline{.568}  &\textbf{.575} & .522& .455 &.500 & .590 & .558 & .537 & .570&\textbf{.631} & \textbf{.675} &.565  \\ 
% \textsc{Instruct} &  & &  &  & &  &  &  & & &  &  & &  &  &  & &   \\ 
\textsc{Reflect} &.543 &.457  & .470 &.566 & .516 & \underline{.588} &{.550}  &{.534}  & .528&.560  & .468&.498  & .498 &.498  & \textbf{.628}&.592&\underline{.592}  &.542 \\ 
\textsc{Moreinfo} &.560 &.611  & .451 &.534 &.300  &.401  & .320 & .390 & .446 & .524 &.517 & .445 & .476 &.403  &.385 &.343 & .386&.435 \\ 
\textsc{Conflict}  &.522 &.534  &.522  & .530 & \underline{.608} &.562  &\textbf{.599}  & \underline{.549}  & \underline{.553} &.567  &.641 &\underline{.594}  &.588  & \textbf{.597} &\underline{.611} & \underline{.597} &.561 &\textbf{.595} \\ 
\textsc{Multi-Related}  & .565 & \textbf{.625}& .520  &.570 &.505  &.490  &.485  &.404  & .521& \underline{.571} &\underline{.645} &.585  &.571 & \underline{.567} &.529 & .455 & .461  & .548 \\ \hline
\textsc{Causal-native}&\underline{.570} & {.590}  & .480  &\textbf{.600} & .493 &.565 & .455&.503   &.532 &.558  &.568  & .585 &\underline{.592} &.424  &.515 &.394&.425 &.508   \\
\textsc{Causal-multi} &\textbf{.575}&\underline{.620}& \underline{.523} & \underline{.595} &.570 &.581  &\underline{.568}  &.500  &\textbf{.567} &\textbf{.593}  &\textbf{.658}& \textbf{.641} &\textbf{.631}  & {.558}  &.547 & .519&.513 &\underline{.583}  \\ \hline
\toprule
\end{tabular}
\caption{Performance comparison of \textit{CausalAbstain} in native language (\textsc{Causal-Native}) and multi-language (\textsc{Causal-Multi}) settings, against the calibration, prompting, consistency, and feedback-based baselines on two datasets. We present the performance of the Abstain Accuracy metric from high-/medium-resource languages (Chinese, Italian, Arabic, and Indonesian) to low-resource languages (Bengali, Telugu, Nepali, and Kannada). $\mathrm{Overall}$ denotes average performance for all languages.
%\textsc{Avg-L} denotes average performance for low-resource languages. 
The best and second-best results are highlighted in \textbf{bold} and \underline{underline}, respectively.}
\label{table:main-results}
\end{table*} 

\section{Evaluation}
\subsection{Experimental Settings}
\paragraph{Datasets \& Languages.} We evaluate \textit{CausalAbstain} using the Hellaswag (M-Hellaswag)~\cite{lai2023okapiinstructiontunedlargelanguage} and Multilingual MMLU (M-MMLU) datasets, which focus on QA using general and commonsense knowledge in multiple languages. Following~\citet{lai2023chatgpt,lai2023okapiinstructiontunedlargelanguage}, we categorize languages based on their data ratios in the pre-training corpus. We randomly sample 500 instances for testing and 200 for validation from high-/medium-/low-resource languages. Additionally, we adopt the related language settings from~\citet{feng-etal-2024-teaching} and set the number of iterations to $N=3$. The analysis of iteration times and language relatedness is provided in Appendix~\ref{app:analysis}.
%\wei{not grammatical nor complete. What do you mean?}, 
 Further details on the datasets and language lists are described in Appendix~\ref{app:experiment_details}. 
\paragraph{LLMs.} We conduct experiments with three LLMs: ChatGPT and GPT-4o (two general black-box LLMs with strong multilingual capabilities~\citep{bang2023multitask,openai2023gpt}); Aya-13B (a multilingual open-source model).
\paragraph{Baselines} We compare \textit{CausalAbstain} against existing approaches that are adaptable to multiple languages: \textit{Calibration-based}: \textsc{Ask Cali}~\citep{tian-etal-2023-just}; 
% \textit{training-based}:  \textsc{Instruct}~\citep{ouyang2022traininglanguagemodelsfollow};
\textit{Prompting-based}:  \textsc{Reflect} and  \textsc{Moreinfo}~\citep{kadavath2022languagemodelsmostlyknow,feng2024knowledgecardfillingllms}; \textit{Consistency-based}: \textsc{Conflict}~\citep{feng-etal-2024-dont}; and \textit{Feedback-based}:  \textsc{Multi-Related}~\citep{feng-etal-2024-teaching}.
\paragraph{Metrics.} We follow the Abstain Accuracy metrics, defined as $\frac{TP+TN}{TP+TN+FP+FN}$ proposed by~\citet{feng-etal-2024-dont}, which evaluates whether the abstention decisions are correct. That is, an LLM should abstain when it would produce an incorrect answer and should not abstain when it would give a correct answer. The term $TP\text{+}TN$ represents cases where the LLM makes correct abstention decisions, including 1) the answer is correct, and the model does not abstain; 2) the answer is incorrect, and the model abstains. 
\paragraph{Answer Likelihood Distribution.} Since LLMs are black-box models~\cite{gat2023faithfulexplanationsblackboxnlp, cheng2024blackboxpromptoptimizationaligning}, directly %calculating causal effect on the parameterization of 
% conducting causal inference on their parameterization is challenging. Instead, we analyze how the final abstention decision $D$ reacts to different $Q$ and $A$ under the influence of feedback $F$. 
conducting causal inference with parameterization of $D$ is challenging. Thus, for each query-answer pair, we repeatedly generate $N$ samples as a representative dataset to approximate distribution $D$. The likelihood distribution $\hat{D}$ can be modeled as a binomial distribution, where the probability of $\hat{D}=1$ is computed using a softmax function applied to the average indicator values, $\bar{I}_1$ and $\bar{I}_0$, of $\hat{D}_j$, representing the decision corresponding to the $j$-th feedback:
$P(\hat{D} = 1) = \frac{\exp\left(\bar{I}_1\right)}{\exp\left(\bar{I}_1\right) + \exp\left(\bar{I}_0\right)}$, where $\bar{I}_1 = \frac{1}{N} \sum_{j=1}^{N} \mathbb{I}(\hat{D}_j = 1),  \bar{I}_0 = \frac{1}{N} \sum_{j=1}^{N} \mathbb{I}(\hat{D}_j = 0)$, and $\mathbb{I}$ is the indicator function, while $N$ is both the number of iterations and the sample size.

\subsection{Experimental Results}
In Table~\ref{table:main-results}, we present the Abstain Accuracy results for three LLMs evaluated on two multilingual datasets, M-MMLU and M-Hellaswag.

\paragraph{\textsc{Causal-multi} achieves state-of-the-art performance.} Our proposed \textsc{Causal-Multi} outperforms the
the strongest baseline in 4 out of 6 settings (across three models and two datasets), achieving an average accuracy improvement of 3.5\% over the best-competing method. Across the 8 different resource languages, \textsc{Causal-Multi} rank the first in 3.3 languages and second in 2.5 languages on average. In contrast, \textsc{Multi-Related} is a method that leverages multiple feedback sources using LLMs without selecting feedback. While it performs well in high- and medium-resource languages with GPT-4o, it struggles in low-resource languages (e.g., Nepali; Kannada) within ChatGPT. 
This performance drop could be attributed to GPT-4o's stronger ability to handle unreliable feedback in high-resource and medium-resource languages, suggesting that merely utilizing LLMs to assess all feedback does not always yield optimal results. 
These findings indicate that leveraging causal effects to filter helpful multilingual feedback can greatly enhance LLMs' performance in different languages, especially in settings where resource languages vary (see \S\ref{section:analysis}).
\setlength{\tabcolsep}{1.3pt} % Default value: 6pt
\renewcommand{\arraystretch}{1.01} % Default value: 1
\begin{table*}[t!]
\center
\footnotesize
\begin{tabular}{lcccccccc>{\columncolor{gray!20}}c|cccccccc>{\columncolor{gray!20}}c} \\\toprule
 & \multicolumn{9}{c}{\textbf{M-MMLU}} & \multicolumn{9}{c}{\textbf{M-Hellaswag}} \\\cline{2-19}
\textbf{Ablative Settings}&\multicolumn{16}{c}{\textbf{\textsc{Aya-13B}}} \\ \hline
 & zh& it & ar& id & bn & te & ne & kn & $\mathrm{Overall}$   & zh& it& ar& id & bn & te & ne & kn & $\mathrm{Overall}$ \\ \hline
\rowcolor{gray!20}
\textsc{Causal-Multi} & .512& .538 & .530 & .535 & .574 & .544 &.545 &.579 &\textbf{.545}  & .588 &.564 & .589 & .626 & .670 &.574 &.597 & .685&\textbf{.612}  \\ 
 \textit{ 1) ignore feedback entirely}  & .465&.470& .530 & .510 &.515 &.551  &.465  &.486 &.499 & .580 & .534& .532 & .545 & .696 & .561& .600 &.571 &{.577} \\ 
 \textit{ 2) consider feedback only} &.512 & .542 & .495 &.500 &.560  &.503  &.540 & .523& .522 & .589 &.564 &.572  & .587 & .670 & .574& .597 &.685 &\underline{.605} \\ 
 \textit{ 3) w/o comparison}  & .491& .542 & .515 &.530 &.555  &.510  &.460  &.555 & .520  &.606  &.572 & .576 & .626 &  .666&.556 & .611 & .685 &\textbf{.612}\\ 
 \textit{ 4) w/o aggregation}  &.577 &.547  & .490 &.500 &.555  &.476  &.574 & .569&\underline{.536}  &.476  &.500 & .502 & .605 & .679 &.583 &.477 &.600 &.553  \\ \hline
 \textbf{}&\multicolumn{16}{c}{\textbf{\textsc{ChatGPT}}} \\ \hline
 \rowcolor{gray!20}
 \textsc{Causal-Multi} &.575&.620& .523 & .595 &.570 &.581  &.568  &.500 &\textbf{.567}  & .593 &.658 & .641 & .631 &.558  &.547 &.519 & .513 &  \textbf{.583} \\ 
 \textit{ 1) ignore feedback entirely}  &.545 &.620  &.505  &.540 &.535  &.510  &.520  &.507 &.536 &.571  &.559 & .593 &.557  & .476 &.457 &.472 &.425 &.514   \\ 
 \textit{ 2) consider feedback only}   & .555&.610  &.495  &.591 &.570  &.565  &.530  &.486 &.550 &  .593 & .658& .598 &.630 &.558  &.547 &.519 & .513& \underline{.577}\\ 
 \textit{ 3) w/o comparison}  &.585 & .613 & .495 &.593 &.565  &.548  &.555  &.493 &\underline{.556}  & .571 &.615 & .580 & .639 &.554 &.502 &.496 & .486& .555 \\ 
 \textit{ 4) w/o aggregation} &.565 &.595  &.500  &.603 &.585  &.561  &.535  &.493 &.555 & .580 &.576 & .563 & .562 &.584 &.524 &.493 & .530& .552 \\ 
\toprule
\end{tabular}
\caption{Ablation study of \textsc{Causal-Multi} on two benchmarks with ChatGPT and Aya-13B. \textsc{Overall} denotes the average performance for all resource languages. The best and second-best results of $\mathrm{Overall}$ are highlighted in \textbf{bold} and \underline{underline}, respectively.}
\label{table:ablative_study}
\end{table*}

\paragraph{Multilingual feedback outperforms monolingual feedback.} We find that \textsc{Causal-native} performs better with stronger LLMs, particularly in high- and medium-resource languages. It achieves the second-best performance with GPT-4o in M-Hellaswag but declines with ChatGPT (dropping from 58.3\% to 50.8\% when compared with \textsc{Causal-Multi}).
%\wei{Not clear. What is compared to Causal-Multi?}. 
This indicates that multilingual feedback may be more beneficial for smaller LLMs. Additionally, the same abstention strategies yield varying results in different languages within the same LLM, which can be attributed to the ratio variations in pre-training data of different languages~\cite{lai2023okapiinstructiontunedlargelanguage}. We further explore the impact of language relatedness of \textsc{Causal-Multi} in Appendix~\ref{app:analysis}.
% within 
% \textsc{Causal-Multi} 
% in Appendix~\ref{app:analysis}.

\subsection{Ablative Study} 
Whether to incorporate feedback in the abstention decision is determined by comparing $\mathrm{NDE}$ and $\mathrm{TIE}^i$. 
To evaluate the effectiveness of this comparison strategy in determining the importance of feedback, we conduct ablation experiments under four different settings:
\textbf{1)} majority voting based on $\hat{D}(Q,A)$ only (ignoring feedback entirely); %w/o any feedback (LLMs make decisions without considering feedback); 
\textbf{2)} majority voting by $\hat{D}(Q,A,\hat{F}^i)$ for all $i$ only (considering feedback only); %w/o NDE (i.e., voting on the final answer across all language feedback); 
\textbf{3)} majority voting by combining both $\hat{D}(Q,A)$ and $D(Q,A,\hat{F}^i)$ (w/o comparison); %w/o filtering (i.e., utilizing NDE and TIE  answers to vote); 
\textbf{4)} majority voting based on $\hat{D}(Q,A,\hat{F}^i)$ only when $\mathrm{NDE}<\mathrm{TIE}^i$ (w/o aggregating over all languages).
%w/o aggregation voted (i.e., directly selecting the answer when $\mathrm{TIE > NDE}$ without aggregation).\wei{The description of the four settings are not clear} The results are presented in Table~\ref{table:ablative_study}.

\paragraph{Multilingual feedback evaluated with the causal framework enhances performance.} When ignoring feedback entirely, the two LLMs demonstrate their poorest performance, with a maximum drop of 9.3\% (from 57.9\% to 48.6\% in Kannada) and an average decrease of 5\% and 4.1\% across both LLMs. When LLMs make the abstention decisions without comparison (i.e., combine both $\hat{D}(Q,A)$ and $\hat{D}(Q,A,\hat{F}^i)$), we observe an average performance drop of 3.7\%, with a maximum decrease of 8\% in Telugu (from 54.5\% to 46\%). The performance of considering feedback only drops maximum of 2.3\% in ChatGPT, suggesting that some feedback may introduce bias and negatively influence the performance. 
\paragraph{Our aggregated voting strategy across related languages outperforms voting solely on language $i$ where $\mathrm{TIE}^i > \mathrm{NDE}$.} While in some languages, such as Nepali (57.4\% on M-MMLU) and Telugu (58.3\% on M-Hellaswag), providing the abstain decision based on $\mathrm{TIE^i} > \mathrm{NDE}$ without aggregation performs slightly better, we observe considerable performance drops in other languages, such as 8.7\% in Arabic (Aya-13B) and 6.9\% in Indonesian (ChatGPT) on M-Hellaswag. This variation is likely due to inconsistent language representation in the pre-training data of LLMs~\cite{lai2023okapiinstructiontunedlargelanguage}. \textsc{Causal-multi} utilizes the aggregated voting mechanism, successfully achieving greater robustness in different languages.
%\wei{you use 'across languages' in many places, but the method is not cross-lingual. So, be careful when using terms}.
We further provide a case where relying solely on the language with $\mathrm{TIE^i > NDE}$ (as seen in the feedback and abstention decision for Dutch) results in an incorrect final abstention decision in Table~\ref{table:case3}.

\begin{table*}[t!]
\center
\tiny
\setlength{\tabcolsep}{5pt} % 默认值是6pt
% 设置行之间的间隙
\renewcommand{\arraystretch}{1}
\begin{tabular}{p{15.8cm}}
\toprule
{\textbf{Question}: 集体安全的含义是什么？A.多个国家共同行动进行自卫的权利 B.盟友为维护受害国家的权利进行自卫的权利 C.经联合国安理会授权使用武装力量 D.经联合国大会授权的维和行动 
(\textcolor{cyan}{Translate}: What is the meaning of collective security? A. The right of several countries to act together in self-defense B. The right of allies to defend themselves in order to protect the rights of the victim country C. The use of armed forces authorized by the UN Security Council D. Peacekeeping operations authorized by the UN General Assembly)} \\
% \hline
\textbf{Proposed Answer}: C.经联合国安理会授权使用武装力量 
(\textcolor{cyan}{Translate}: C. The use of armed forces authorized by the UN Security Council) \\
\hline
\textbf{No Feedback}: \textit{Please review the correctness of proposed answer True or False directly.}.
Iterate 1: \textcolor{c1}{True}, Iterate 2: \textcolor{c1}{True}, Iterate 3: \textcolor{c1}{True}. --> \textcolor{c2}{NDE = 0.0285}\\
\hline
\textbf{Feedback in native language}: \textit{Please review the proposed answer and provide a paragraph of feedback on its correctness. Feedback should be in Chinese.}\\
\setlength{\tabcolsep}{5pt} % 默认值是6pt
% 设置行之间的间隙
\renewcommand{\arraystretch}{1}
\begin{tabular}{@{}p{5.1cm}|  p{5.5cm} | p{4.6cm}}
\textbf{Iterate 1}:
你选择的答案C，并不完全符合“集体安全”的定义...”...
(\textcolor{cyan}{Translate}: False. The answer you chose, C does not fully meet the definition of "collective security". The concept of collective security usually refers to ... Therefore, the most appropriate answer is A, which means "the right of multiple countries to act together in self-defense"...) \textcolor{red}{\XSolidBrush} & \textbf{Iterate 2}:
True。答案的确是C...
(\textcolor{cyan}{Translate}: True. The answer is indeed C. The meaning of collective security refers to ensuring the security of all countries in the international system through multilateral cooperation and coordination with international organizations (such as the United Nations). If a country is invaded or threatened... Therefore, option C is correct...) \textcolor{green}{\Checkmark} & \textbf{Iterate 3}:
所提出的答案C是正确的...
(\textcolor{cyan}{Translate}: True, proposed answer C is correct .. The UN Security Council plays a central role in collective security mechanisms, particularly in authorizing the use of armed force. Therefore, option C accurately reflects the core mechanism of collective security...) \textcolor{green}{\Checkmark} \\
    % \hline
    \end{tabular}
% } \\
\\
\hline
Iterate 1: \textcolor{c1}{False}, Iterate 2: \textcolor{c1}{True}, Iterate 3: \textcolor{c1}{True}.  
 --> \textcolor{c2}{TIE = 0.0123} \\
\textbf{CausalAbstain:} NDE > TIE, then do not utilize feedback. The final causal abstain decision voted by ``No Feedback"  is: \textcolor{c3}{\textbf{Do Not Abstain}}. (The correct answer is C.) \\
\hline\hline
\textbf{Question}: 对于复数 $z = a + bi$，已知 a 和 b 为正整数，并且 $z^3 + z^2 + z$ 是实数。找出最小可能的 b 值。A. 10 B.1 C.6 D.11
(\textcolor{cyan}{Translate}: For a complex number $z = a + bi$, given that a and b are positive integers, and $z^3 + z^2 + z$ is a real number. Find the smallest possible value of b. A. 10 B.1 C.6 D.11)\\
\textbf{Proposed Answer}: C.6\\\hline
\textbf{No Feedback}: \textit{Please review the correctness of proposed answer True or False directly.}\\
Iterate 1: \textcolor{c1}{True}, Iterate 2: \textcolor{c1}{False}, Iterate 3: \textcolor{c1}{True}. --> \textcolor{c2}{NDE = 0.0034}\\\hline
\textbf{Feedback in native language}: \textit{Please review the proposed answer and provide a paragraph of feedback on its correctness. Feedback should be in Chinese.}\\
% \multicolumn{1}{p{15.5cm}}{%
    % \setlength{\tabcolsep}{5pt} % 默认值是6pt
    % 设置行之间的间隙
    % \renewcommand{\arraystretch}{1.5}
\begin{tabular}{@{}p{5.6cm}| p{4.9cm}| p{4.6cm}}
    % \hline
\textbf{Iterate 1}:
False. 对于复数$z=a+bi$...选项 D 是正确的，而提议的答案 C 是不正确的。
(\textcolor{cyan}{Translate}: False. For the complex number $z=a+bi$, it is known that $a$ and $b$ are positive integers, and $z3+z2+z$ is a real number. We need to find the smallest $b$ value that satisfies the condition. From the above derivation... When $a=6$, b=11 is an integer. Therefore, the minimum b value is 11. So option D is correct and proposed answer C is incorrect.) \textcolor{green}{\Checkmark} & \textbf{Iterate 2}:
False。你的答案是 C，但我们需要验证它是否正确...答案 C 是不正确的。
(\textcolor{cyan}{Translate}: Your answer is C, but we need to verify whether it is correct. We consider the complex number $z=a+bi$, where a and b are positive integers, and $z^3+z^2+z$ is a real number... Therefore, answer C is incorrect. Please recheck the calculation process, the correct answer should be b = 1) \textcolor{green}{\Checkmark} & \textbf{Iterate 3}:
你所提出的答案 C (6) 是正确的...
(\textcolor{cyan}{Translate}: Your answer C (6) is correct. To verify this, we need to consider the condition that the complex number $z=a+bi$, that is, $z^3+z^2+z$ must be a real number. After calculation and analysis, we find that this condition can be satisfied when b=6. Therefore, option C is the correct answer. Your answer is correct...) \textcolor{red}{\XSolidBrush} \\
    % \hline
\end{tabular}
 \\\hline
Iterate 1: \textcolor{c1}{False}, Iterate 2: \textcolor{c1}{False}, Iterate 3: \textcolor{c1}{True}. 
 --> \textcolor{c2}{TIE = 0.0137}\\
\textbf{CausalAbstain:} NDE < TIE, then choose to use the feedback. The final causal abstain decision voted by ``Feedback in native language" is: \textcolor{c3}{\textbf{Abstain}}. (The correct answer is D.)\\
\toprule
\end{tabular}
\caption{Two examples in the native language (Chinese). In the first one, \textit{CausalAbstain} chooses not to use feedback and answers correctly. In the second one, it effectively decides to abstain proposed answer with feedback, while the LLM gives an incorrect answer without feedback. We condense the Chinese expressions and primarily showcase the English translations (\textcolor{cyan}{Translate:}) for readability. The full context is provided in Table~\ref{table:case2} and~\ref{table:case1}.}
\label{table:case12}
\end{table*}

\section{Analysis}\label{section:analysis}
\paragraph{Case Study.} We represent \textbf{three types of cases}: \textbf{1)} Without feedback (Table~\ref{table:case12}): LLMs can make the correct abstention decision without any feedback. However, biased feedback may negatively impact the final decision; \textbf{2)} Feedback-driven (Table~\ref{table:case12}): Feedback plays a crucial role in influencing the abstention decision and helps the model achieve correct conclusion;  \textbf{3)} Multilingual feedback scenarios (Table~\ref{table:case3}): Helpful feedback from related languages leads to the correct abstention decision, whereas relying solely on feedback in the native language proves ineffective.

\setlength{\tabcolsep}{0.5pt} % Default value: 6pt
\renewcommand{\arraystretch}{1.15} % Default value: 1
\begin{table}[ht!]
\center
\scriptsize
\begin{tabular}{lcc|cc|cc|cc|cc|cc} \\\toprule
& \multicolumn{6}{c|}{\textbf{M-MMLU}} & \multicolumn{6}{c}{\textbf{M-Hellaswag}} \\\cline{2-13}
\multirow{2}{*}{Method}& \multicolumn{2}{c|}{\tiny{Aya}} & \multicolumn{2}{c|}{\tiny{ChatGPT}} & \multicolumn{2}{c|}{\tiny{GPT-4o}}& \multicolumn{2}{c|}{\tiny{Aya}} & \multicolumn{2}{c|}{\tiny{ChatGPT}} & \multicolumn{2}{c}{\tiny{GPT-4o}}\\
& H/M & L& H/M & L& H/M & L & H/M & L& H/M & L& H/M & L \\ \hline
\textsc{Ask Cali} &\textbf{.557} &.507&.447 &\textbf{.596} &.229 &.456  &.519& .503 &.526&\textbf{.603}&.293&.439 \\
\textsc{Reflect} &.471 &.376& .509&.547&.521 &.487 & .488 &.487& .506&{.578}&.321&.459\\
\textsc{Moreinfo} &.479 &.385&.539 &.353& .766& .606& .500 &.485 &.491&.379&.663&.524\\
\textsc{Conflict} &.484 &\underline{.524}& .527&.580& .679& .519 &.512&.488 &\underline{.598}&\underline{.592}&.453&.589\\
\textsc{Multi-Related} &\underline{.538} &\textbf{.561}&\underline{.570} &.471&\textbf{.790} &\underline{.660}  &.500&.556 &.593&.503&.768&.692\\\hline
\rowcolor{gray!20}
\textsc{Causal-Native} & .501&.507& .560&.504& .782&.644  &\underline{.580}&\underline{.586} &.576&.440&\underline{.789}&\underline{.710}\\
\rowcolor{gray!20}
\textsc{Causal-Multi} &.529 &\textbf{.561}&\textbf{.578} &\underline{.555}&\underline{.788} & \textbf{.687} &\textbf{.592}&\textbf{.632} &\textbf{.631}&.534&\textbf{.792}&\textbf{.716}\\
\toprule
\end{tabular}
\caption{The average performance of high-/medium-resource (H/M) and low-resource languages (L).}
\label{table: high/medium/low}
\end{table}

\begin{figure}[ht!]
    \centering
    \includegraphics[width=0.5\textwidth]{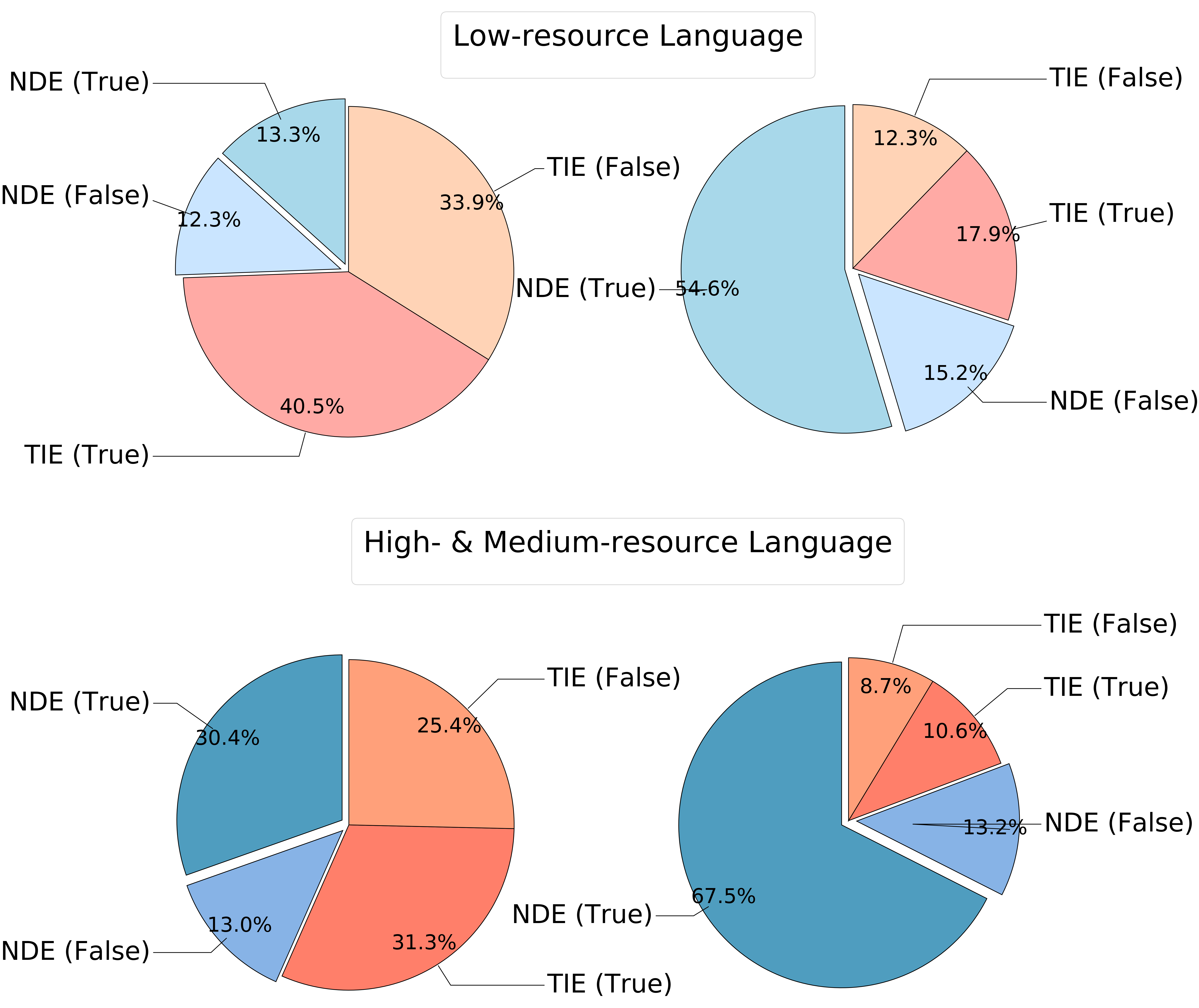}
    % \vspace{-0.8m}
    \caption{The distribution and accuracy rate of NDE and TIE with ChatGPT (left column) and GPT-4o (right column) in different resource languages.}
    % \vspace{-0.6cm}
    \label{fig: pie}
    \label{fig:enter-label}
\end{figure}

\paragraph{Performance in different resource languages.} In Figure~\ref{fig: pie}, we illustrate the distribution of NDE and TIE in low-, high- and medium-resource languages with ChatGPT and GPT-4o. We can see from the results of the left and right columns (ChatGPT vs. GPT-4o), GPT-4o exhibits a higher proportion of NDE and a lower proportion of TIE. This suggests that more powerful LLMs tend to have higher confidence in their proposed answers, leading to greater consistency when reviewing their own responses. Furthermore, when comparing the top and bottom rows (language resource levels), both LLMs show a higher proportion of NDE in high- and medium-resource languages, indicating greater confidence in their answers for well-represented languages. Additionally, Table~\ref{table: high/medium/low} reveals that the average performance gap between different resource languages is smaller in Aya-13B. This suggests that Aya-13B, being explicitly multilingual, performs more consistently among various languages compared to the general-purpose GPT-4o and ChatGPT.

\paragraph{\textsc{Causal-Random} and \textsc{Causal-English}.} We employ multiple languages for feedback generation, but the languages are randomly selected from a language pool, which is named \textsc{Causal-Random}. As shown in Table~\ref{table:more_results_Causal-Random}, we observe that overall, \textsc{Causal-Multi} performs better than \textsc{Causal-Random} and \textsc{Causal-Native}, which showcase the importance of choosing the related languages. This implies that the source of performance improvement comes from both the languages selected and the CausalAbstain method.
As shown in Table~\ref{table:more_results_Causal-Random}, in GPT-4o, \textsc{Causal-English} underperforms \textsc{Causal-Native}. In ChatGPT, the overall performance of \textsc{Causal-English} is better than \textsc{Causal-Native}, which is similar to the findings in previous research~\cite{shi2022language,zhang2023don} that indicate ChatGPT performs better in English than other languages. 

\setlength{\tabcolsep}{1pt} % Default value: 6pt
\renewcommand{\arraystretch}{1.05}
\begin{table}[t!]
\center
\scriptsize
\begin{tabular}{lcccccccc>{\columncolor{gray!20}}cc} \\\toprule
&\multicolumn{10}{c}{\textsc{Causal-Random}} \\ \hline
\textbf{LLMs} & zh& it & ar& id & bn & te & ne & kn & Overall & \makecell[c]{Overall\\ (\textsc{Causal-Multi})}\\ \hline
ChatGPT	&.584	&.615&	.598&	.596	&.571&	.524&	.493&	.513&	.562 &	.583\\ 
GPT-4o	&.746	&.807	&.710&	.780	&.714	&.653	&.748	&.621	&.722 &	.754\\ 
Aya-13B	&.525	&.508&	.510	&.600	&.666	&.520	&.582	&.647	&.570	& .612\\ \hline\hline
% \textbf{LLMs}&\multicolumn{10}{c}{\textbf{{M-Hellaswag}}} \\ \hline
&\multicolumn{10}{c}{\textsc{Causal-English}} \\ \hline
\textbf{LLMs} & zh& it & ar& id & bn & te & ne & kn & Overall & \makecell[c]{Overall \\(\textsc{Causal-Native})}\\ \hline
ChatGPT	&.571	&.602&.681	&.570	&.588&	.601&.532&.561&.588&.508\\ 
GPT-4o	&.785	&.794	&.690	&.800&.688&.666	&.716&	.644&	.723	&.749\\ 
Aya-13B	&.549&.559&.515	&.575&.675&.515	&.611&	.616&	.577	&.564\\ 
\toprule
\end{tabular}
\caption{The results of \textsc{Causal-Random} and \textsc{Causal-English} in M-Hellaswag. The overall performance of \textsc{Casual-Multi} and \textsc{Causal-Native} is presented for comparison purposes.}
% \vspace{-0.6cm}
\label{table:more_results_Causal-Random}
\end{table}

\section{Related Work}
\paragraph{Causal for LLMs} 
%With the advent of generative LLMs, attracting increased research interest in applying causal inference to address LLM-related challenges and augment their functionality. 
LLMs can significantly benefit from causality, as it enhances their ability to understand and reason about cause-and-effect relationships within data~\cite{liu2024largelanguagemodelscausal}. A large body of research has explored this~\cite{li2021causalbertinjectingcausalknowledge,jin2024cladderassessingcausalreasoning,wang2023colacontextualizedcommonsensecausal,jin2024largelanguagemodelsinfer,jiang2024llm4causaldemocratizedcausaltools,Chen_2024}, including identifying knowledge bias pre-trained in LLMs that can lead to incorrect answers and hallucinations~\cite{zhang2024knowledgealignmentproblembridging, wu-etal-2024-decot}; improving LLMs performance in specific tasks, such as visual question answering~\cite{zhao2023causalcogcausaleffectlookcontext, zang2023discovering}.

\paragraph{Abstention.} Existing abstention strategies can be categorized as follows: \textit{Calibration-based} strategies aim to extract confidence score to gauge uncertainty~\cite{tian-etal-2023-just,kuhn2023semanticuncertaintylinguisticinvariances}. \textit{Prompt-based} strategies employ instructions to induce reflection and determine whether the generated answer is reliable~\cite{kadavath2022languagemodelsmostlyknow,edunov2018understandingbacktranslationscale,si2023gettingmixturelanguagemodel}. 

\textit{Consistency-based} and \textit{collaboration-based} strategies nvolve multiple LLMs reflecting on their answers cooperatively~\cite{feng-etal-2024-dont,wang2023selfconsistencyimproveschainthought}. Recent \textit{feedback-based} methods adapt to multilingual settings. However, these approaches either underperform outside of English or struggle with generated feedback quality. To address this gap, we propose a causal method that identifies helpful feedback, improving multilingual abstention.
\section{Conclusion and Future Work}
We propose a novel causal method, \textit{CausalAbstain}, to assist LLMs in abstaining from providing incorrect answers. Extensive experiments on two datasets demonstrate that our method, \textsc{Causal-Multi}, achieves state-of-the-art performance in multilingual settings, while \textsc{Causal-native} exhibits competitive performance in monolingual settings, suggesting the effectiveness of abstention based on causality. 
In future work, we will explore additional contextual factors, such as model uncertainty and external knowledge sources, to further improve abstention decisions across diverse tasks and languages.
\section*{Limitation}
Our approach, \textsc{Causal-multi}, leverages multilingual feedback to help LLMs make abstention decisions. While it requires prompt LLMs multiple times, leading to higher inference costs compared to simpler prompting approaches, it is still not the most computationally expensive method compared with~\cite{feng-etal-2024-dont}. To mitigate the cost, we introduce \textsc{Causal-Native}, which reduces the number of prompting requests while maintaining competitive performance. We compare the LLMs' overhead during inference using different abstention strategies, shown in Table~\ref{table:overhead}.

Furthermore, our causal graph in Figure~\ref{fig: causal graph} provides a general framework for formulating the abstention task. Future work could explore incorporating latent variables, such as LLM biases, to further refine the approach.

\section*{Ethics Statement}
Our study on mitigating bias and hallucinations in LLMs acknowledges the ethical implications of data-driven biases in AI, particularly their impact on performance. All experiments were conducted using publicly available datasets, and no human participants were involved.

\section*{Acknowledgement}
This work is partially supported by Tencent Rhino-Bird Focused Research Program (Value-aligned Credible Large Language Model) and RMGS project (Artificial Intelligence and Big Data Analytics for Social Good).

% \input{latex/Ethical Statement}
% \input{2-related_work}
% \input{3-method}
% \input{4-experiment}
% \input{5-Discussion}
% \input{6-conclusion}
% % \input{6-limitation}
% \input{7-ethical}

\bibliography{custom}
\appendix
\newpage
\section*{\Large Appendix}

\setlength{\tabcolsep}{3pt} % Default value: 6pt
\renewcommand{\arraystretch}{0.92}
\begin{table*}[htbp!]
\center
\footnotesize
\begin{tabular}{l>{\columncolor{gray!20}}ccccccccc>{\columncolor{gray!20}}c>{\columncolor{gray!20}}c} \\\toprule
 & \multicolumn{8}{c}{\textbf{GPT-4o}}  \\\hline
&\multicolumn{8}{c}{\textbf{\textsc{M-MMLU}}} \\ \hline
\textbf{Ablative Settings}& Avg-HM & zh& it & ar& id & bn & te & ne & kn & Avg-L & Overall \\ \hline
\rowcolor{gray!20}
\textsc{Causal-Multi} &.788&.765 &.830 &.760	&.795	&.740	&.670	&.675	&.671 &.686 &.738 \\ 
 \textit{ 1) ignoring feedback entirely}&.761 &.755& .830&.740	&.720	&.670	&.531	&.595	&.636 &.608 & .685\\ 
 \textit{ 2) considering feedback only} &.781 &.760& .815&.765	&.785	&.725	&.639&	.685	&.630 &.670 & .726\\ 
 \textit{ 3) w/o comparison}& .786&.760&.825 &.760	&.800&	.735	&.578&	.655	&.623& .648 &.717\\ 
 \textit{ 4) w/o aggregating across all languages}& .783&.755& .825&.760	&.790	&.715&	.660	&.690&	.630 &.674 &.728\\ \hline
 & \multicolumn{8}{c}{\textbf{M-Hellaswag}} \\\hline
 & Avg-HM & zh& it& ar& id & bn & te & ne & kn & Avg-L & Overall\\
 \rowcolor{gray!20}
\textsc{Causal-Multi} &.620 &.593 & .658 &.598&.630 &.558&.547&.519	&.513&.534&.577\\ 
  \textit{ 1) ignoring feedback entirely} & .570&.571 &.559&.593&.557& .476 &.457 &.472&.425&.458&.514\\ 
 \textit{ 2) considering feedback only} & .620&.593 & .658 &.593&.630 &.558&.547	&.519	&.513&.534&.576 \\ 
 \textit{ 3) w/o comparison} &.601& .571 & .615 &.580&.639& .554	&.502	&.497&.486&.510&.556\\ 
 \textit{ 4) w/o aggregating across all languages} &.570& .580 &.576 & .563&.562 &.584	&.524&	.493 &.530&.533&.552\\
\toprule
\end{tabular}
\caption{Ablation study of \textsc{Causal-Multi} on two benchmarks using GPT-4o. The results include the average performance for high- and medium-resource languages (Avg-HM), low-resource languages (Avg-L), and overall performance across all languages.}
\label{table:GPT-4o_ablative}
\end{table*}

\setlength{\tabcolsep}{2pt} % Default value: 6pt
\renewcommand{\arraystretch}{1.02}
\begin{table*}[ht]
\center
\footnotesize
\begin{tabular}{lcccccccc>{\columncolor{gray!20}}c|cccccccc>{\columncolor{gray!20}}c} \\\toprule
 & \multicolumn{9}{c}{\textbf{LLaMa}} & \multicolumn{9}{c}{\textbf{Phi}} \\\hline
\textbf{Method}&\multicolumn{18}{c}{\textbf{\textsc{M-MMLU}}} \\ \hline
& zh& it & ar& id & bn & te & ne & kn & Overall&zh& it & ar& id & bn & te & ne & kn & Overall \\ \hline

\textsc{Ask Cali}&\textbf{.646}&	.442&	\textbf{.695}	&\textbf{.662}&	.269	&.388&	.304&	.473&.485 &.381&	.423 &.490&	.435&	.442&	.442&	.514&	.467&	.449 \\ 
\textsc{Reflect}&	.419&	.519&	.570&	\underline{.546} &	\textbf{.531}&	.504&	.480&	.527&	.512&\textbf{.652}&	.603	&{.570}&	\underline{.578}	&.455	&.450	&.480	&.457	&\underline{.531}\\
\textsc{Moreinfo}&	.536&	.506&	.490&	.474&	\underline{.497} &	.380&	.514&	.473	&.484&.484	&.468	&{.543}	&.487	&.407&	\textbf{.543}&	\textbf{.601}&	.504&	.505\\
\textsc{Conflict}&	.510& \underline{.532}&	.536&	.494	&.421	&\underline{.527}&	.514	&.434&	.496&.568&	.526	&.457&	.435	&.476	& \underline{.504} &	.527&	.527	&.503\\
\textsc{Multi-Related} &	.497&	\underline{.532}&	.477	&.481&	.476	&.466&	.460	&.442&	.479&{.497}	&.532	&.563&	.455	&\textbf{.497}	&.442	&\underline{.581}	& \textbf{.543}	&{.514}\\\hline
\textsc{Causal-Native} &	.574	&.474&	.536	&.532&	.455&	.480&	\underline{.580}&	\underline{.542} &\underline{.522}&{.516}	&\underline{.608}&	\underline{.589}	&\textbf{.584}	&.441&	.410	&.445	&.500&	.512\\
\textsc{Causal-Multi} &	\underline{.612}&	\textbf{.589}	&\underline{.576}	&.523	&.496&	\textbf{.581}&	\textbf{.581}&	\textbf{.558}&	\textbf{.565} &\underline{.574}&	\textbf{.647}&	\textbf{.609}	&\textbf{.584}	& \underline{.482} &	.441	&.506&	\underline{.531}	&\textbf{.547}\\\hline
\toprule
\end{tabular}
\caption{The abstention accuracy in M-MMLU in LLaMa and Phi, the \textbf{bold} and \underline{underline} represent the best and second performance, respectively.}
\label{table:additional_results}
\end{table*}

\section{Analysis}\label{app:analysis}
\paragraph{The ablative study of GPT-4o.} We present the ablation study of GPT-4o, as shown in Table~\ref{table:GPT-4o_ablative}. 
Similar trends are observed in the ablation experiments on ChatGPT and Aya-13B. GPT-4o demonstrates the poorest average performance when feedback is entirely ignored, with a drop of 7.6\% on Avg-L (from 53.4\% to 45.8\% in M-Hwllaswag). When GPT-4o makes the abstention decisions without comparison (i.e., combining both $\hat{D}(Q,A)$ and $\hat{D}(Q,A,\hat{F}^i)$), we observe an average performance drop of 3.8\% in M-MMLU and 2.4\% in M-Hellaswag on Avg-L. Additionally, the absence of feedback results in relatively good performance for relatively high-resource languages, aligning with the observations in Figure~\ref{fig: pie}.

\begin{figure}[htbp!]
    \begin{minipage}[t]{0.23\textwidth}
        \centering
        \includegraphics[width=1\textwidth]{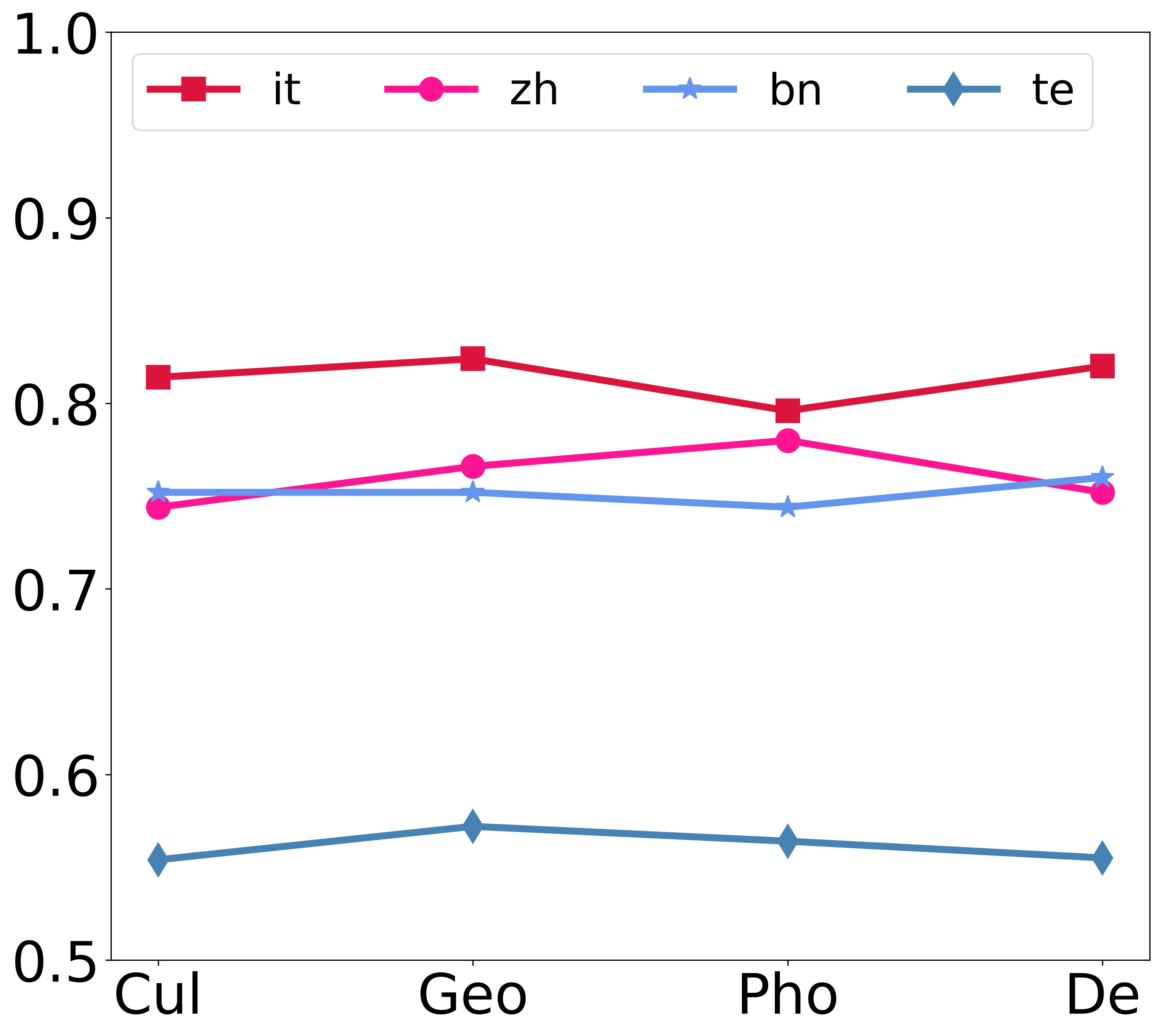}
        % \caption{Log-Transformed Regression}
    \end{minipage}% 
    % \vspace{0.5cm}
    \begin{minipage}[t]{0.24\textwidth}
        \centering
        \includegraphics[width=1\textwidth]{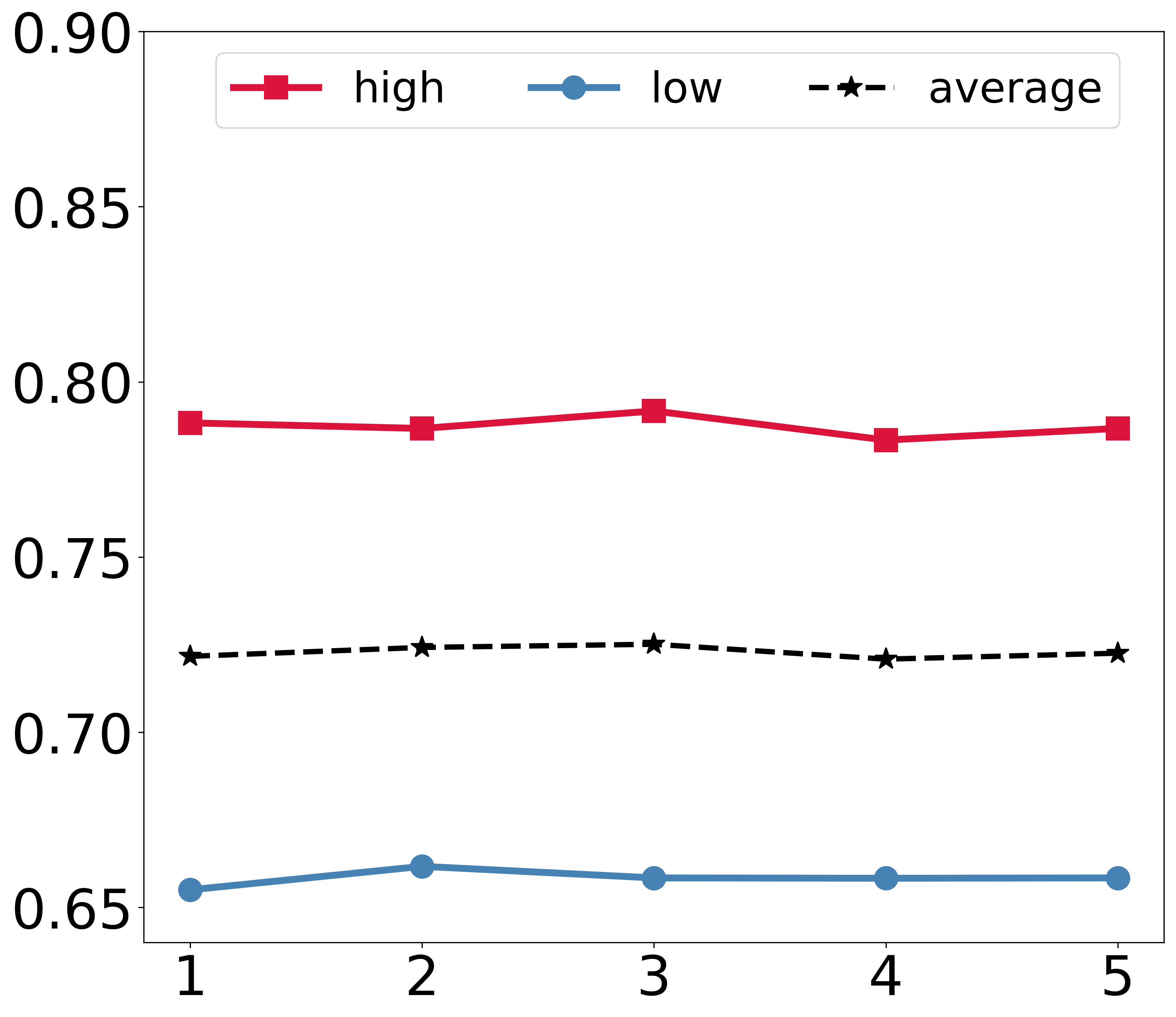}
        % \caption{logistic Regression}
    \end{minipage}
    \caption{The impact of language relatedness (different settings) on specific languages, and the effect of iteration count on high- and low-resource languages.}
    \label{fig: any}
\end{figure}

\paragraph{The influence of language relatedness and iteration count.} Following the six linguistic attributes identified in Lang2vec (i.e., syntactic, geographic, phonological, genetic, inventory, and featural~\cite{littell-etal-2017-uriel}), \citet{feng-etal-2024-teaching} analyzed language relationships based on these attributes. They also introduced the \textit{culture} attribute, which defines related languages within the same cultural cluster according to the World Values Survey. Their findings indicate that \textit{geography}, \textit{phonology} and \textit{culture} are the most influential attributes. Following \cite{feng-etal-2024-teaching}, the related languages are shown in Table~\ref{table:related_languages}. To assess the impact of related languages on \textsc{Causal-Multi}, we conducted an experiment comparing four different settings (culture, geography, phonology, default) as shown in Figure~\ref{fig: any}. Our observations revealed that performance varied across different language queries depending on the choice of related languages. Consequently, we utilized 100 held-out sets in our experiments to determine the most suitable related languages for each specific language.

\begin{figure}[htbp!]
    \begin{minipage}[t]{0.5\textwidth}
        \centering
        \includegraphics[width=0.95\textwidth]{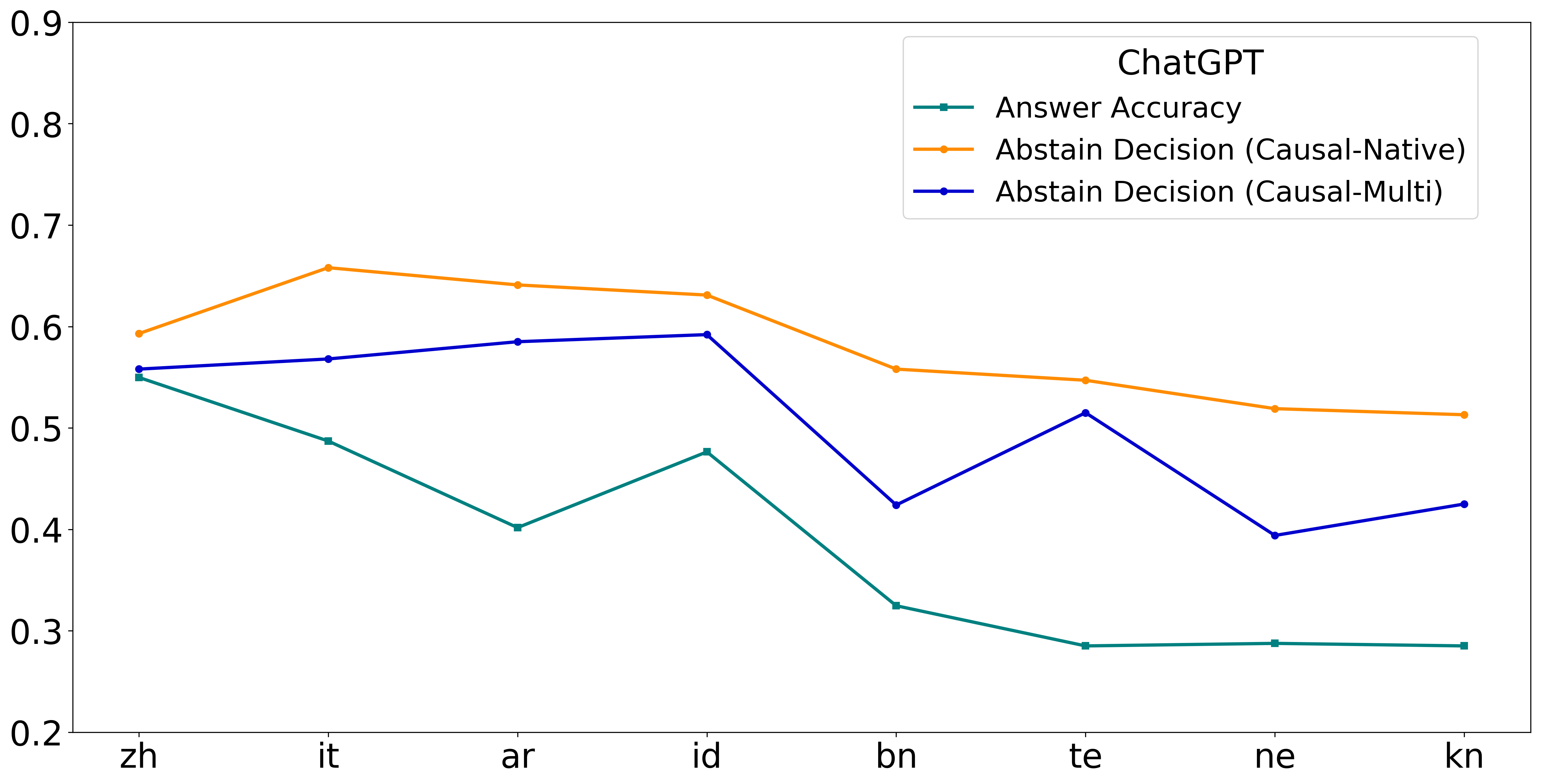}
        % \caption{Log-Transformed Regression}
    \end{minipage}% 

    \begin{minipage}[t]{0.5\textwidth}
        \centering
        \includegraphics[width=0.95\textwidth]{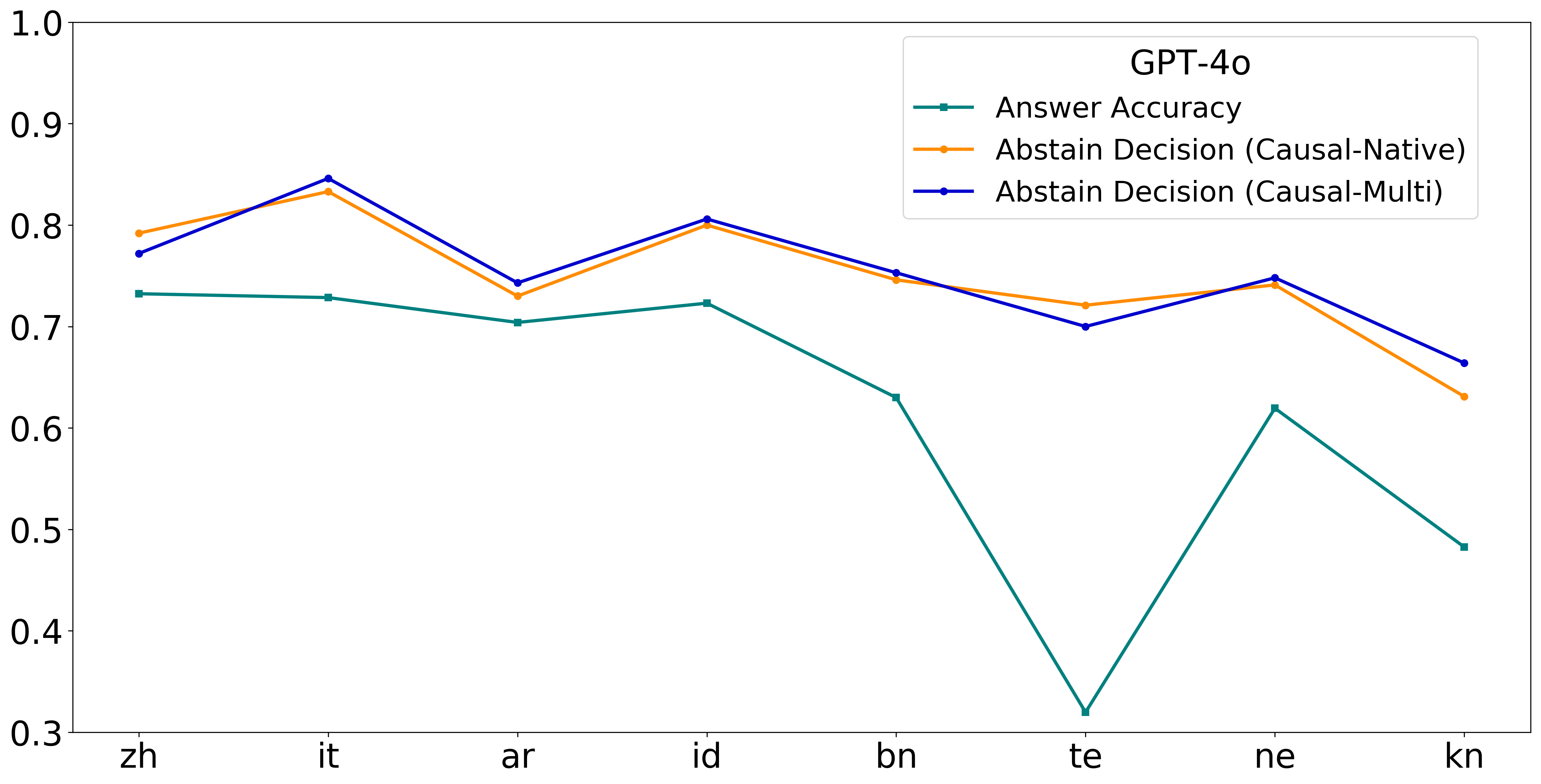}
        % \caption{logistic Regression}
    \end{minipage}
    \caption{Comparison of the accuracy of LLMs' proposed answers and their abstain decisions.}
\label{fig: answer_abstain}
\end{figure}

Additionally, we conduct experiments on the different settings of iteration time from 1-5 with a held-out set (100 instances in the validation set), as shown in Figure~\ref{fig: any}. While variations in the number of iterations have a minimal impact on performance, we have set the iteration count to three, which may yield a slight improvement on average.
\paragraph{The comparison between LLMs' answer accuracy and the abstention accuracy.} In Figure~\ref{fig: answer_abstain}, we observe that there is no synchronized correlation between the two performance metrics. This suggests that abstention is a distinct research question, similar to the observation made by ~\cite{feng-etal-2024-teaching} in other abstention strategy settings.

% \section{More Results}\label{app: more_results}
\paragraph{Results on LLaMa and Phi.} We provide additional results within LLaMa (LLaMa3.2) and Phi (Phi4). As shown in the following Table~\ref{table:additional_results}, we can observe that \textsc{Causal-Multi} outperforms all the baselines with LLaMa and Phi. Meanwhile, \textsc{Causal-Native} shows the second-best performance within LLaMa. 

\begin{table*}[htbp!]
\center
\small
\begin{tabular}{p{16cm}}\toprule
Related language settings:{ "zh": ["Chinese", "Chinese", "Chinese", "English", "Russian", "German","Italian", "Dutch", "Arabic", "Arabic", "Slovak", "Danish"], "it": ["French", "Slovak", "Hungarian","German", "French", "Hungarian","Chinese", "Dutch", "Arabic", "Catalan", "Romanian", "Ukrainian"],  "id": ["Indonesian", "Indonesian", "Indonesian", "Vietnamese",
"Bengali", "Tamil", "English", "Russian", "Catalan","Vietnamese",
"Catalan", "Russian"], "ar": ["Arabic", "Hindi", "Bengali","English", "Russian", "German", "Chinese", "Italian", "Dutch", "Chinese", "Slovak", "Danish"], "bn": ["Arabic", "Hindi", "Bengali","Nepali", "Vietnamese", "Hindi","Telugu", "Kannada", "Russian","Hindi", "Telugu", "Nepali"], "ne":
["Arabic", "Hindi", "Bengali", "Hindi", "Bengali", "Vietnamese", "Kanaada", "Telugu", "Hindi", "Romanian", "Telugu", "Kannada"], "te": ["Arabic", "Hindi", "Bengali", "Hindi", "Tamil", "Malayalam", "Kannada", "Russian", "Catalan", "Kannada", "Tamil", "Nepali"], "kn": ["Arabic", "Hindi", "Bengali", "Tamil",
"Malayalam", "Marathi","Kannada", "Russian", "Catalan", "Telugu", "Malayalam", "Tamil"]}\\
\toprule
\end{tabular}
\caption{Language settings.}
\label{table:related_languages}
\end{table*}

\section{Experiment Details}\label{app:experiment_details}
\paragraph{Dateset.} We use M-MMLU and MHellaswag~\cite{lai2023okapiinstructiontunedlargelanguage} and follow~\citet{lai2023chatgpt,lai2023okapiinstructiontunedlargelanguage} in categorizing languages based on their data ratios in the pre-training corpus. A language is classified as high-resource (HRL) if its data ratio exceeds 1\%, medium-resource (MRL) if it falls between 0.1\% and 1\%, and low-resource (LRL) if it is below 0.1\%. Specifically, we include languages spanning from low to high resource levels: Bengali (bn), Telugu (te), Nepali (ne), Kannada (kn), Arabic (ar), Indonesian (id), Chinese (zh), and Italian (it).

\paragraph{Model details.} We use the Azure OpenAI API checkpoint "gpt-4o-2024-08-06" for GPT-4o, and the "GPT-3.5-Turbo" model checkpoint from OpenAI for ChatGPT. Additionally, we utilize the "CohereForAI/aya-101" model checkpoint from Huggingface for Aya-13B. Following~\citet{feng-etal-2024-dont}, we conduct the experiments on the baselines. 
\setlength{\tabcolsep}{6pt} % Default value: 6pt
\renewcommand{\arraystretch}{1} % Default value: 1
\begin{table}[ht!]
\center
\small
\begin{tabular}{l|c} \toprule
Method &  \# LLM inference requests  \\  \hline
\textsc{Moreinfo} & 2   \\  
\textsc{Ask Cali} & 6   \\ 
\textsc{Reflect} & 2   \\ 
\textsc{Multi-related} & 5 \\
\textsc{Coop-self} & 8 \\\hline
\textsc{Causal-native} & 4 \\
\textsc{Causal-multi} & 10 \\
\toprule
\end{tabular}
\caption{The \# of LLM inference requests for each query with different abstention strategies.}
% \wei{what is the number column?}}
\label{table:overhead}
\end{table}

\setlength{\tabcolsep}{6pt} % Default value: 6pt
\renewcommand{\arraystretch}{1} % Default value: 1
\begin{table}[ht!]
\center
\small
\begin{tabular}{l|c|c} \toprule
$N$ &  \makecell{\# LLM inference \\requests of \\\textsc{Causal-Native}} & \makecell{\# LLM inference\\ requests of \\\textsc{Causal-Multi}}  \\  \hline
1	&2	&4\\\hline
2	&3&	7\\\hline
3	&4&	10\\\hline
4	&5	&13\\\hline
5	&6	&16\\\hline
% \hline
\end{tabular}
\caption{The \# of LLM inference requests for each query with \textsc{Causal-Native} and \textsc{Causal-Multi}, where $N$ is the number of iterations.}
% \wei{what is the number column?}}
\label{table:Inference_time}
\end{table}

\paragraph{LLMs overhead with different abstention strategies.} 
For efficiency, as shown in the following Table~\ref{table:Inference_time}, the computational efficiency of the proposed \textit{CausalAbtsain} is $\mathcal{O}(N)$, where $N$ is the number of iterations. Larger $N$ may lead to higher costs. In addition, various abstention mechanisms have differing computational overheads, with the most significant cost being the number of LLM inference requests for each query. A comparison of these mechanisms, including ours, is provided in Table~\ref{table:overhead}, following the approach outlined by \citet{feng-etal-2024-dont}.

\begin{table*}[htbp!]
\center
\small
% \begin{tabular}{l} \\\toprule
\begin{tabular}{p{16cm}}\toprule
\textbf{Question}: 对于复数 $z = a + bi$，已知 a 和 b 为正整数，并且 $z^3 + z^2 + z$ 是实数。找出最小可能的 b 值。A. 10 B.1 C.6 D.11
(\textcolor{cyan}{Translate}: For a complex number $z = a + bi$, given that a and b are positive integers, and $z^3 + z^2 + z$ is a real number. Find the smallest possible value of b. A. 10 B.1 C.6 D.11)\\
\textbf{Proposed Answer}: C.6\\\hline
\textbf{No Feedback}: \textit{Please review the proposed answer True or False directly.}\\
Iterate 1: \textcolor{c1}{True}, \\Iterate 2: \textcolor{c1}{False}, \\Iterate 3: \textcolor{c1}{True}. --> \textcolor{c2}{NDE = 0.0034}\\\\
\textbf{Feedback in native language}:\\
\textit{Please review the proposed answer and provide a paragraph of feedback on its correctness. Feedback should be in language i.}\\\\
\textbf{Iterate 1}:
False. 对于复数$z=a+bi$，已知$a$和$b$为正整数，并且$z3+z2+z$是实数。我们需要找到满足条件的最小$b$值。从上面的推导，我们有虚部为零的条件：$b(1+2a+3a2−b2)=0$因为$b$为正整数，所以：$1+2a+3a2=b2$我们尝试不同的$a$值以找到最小的$b$值：$a=1$时：... $a=2$时：... $a=6$时，b=11是整数。因此，最小的b值为 11。所以选项 D 是正确的，而提议的答案 C 是不正确的。
(\textcolor{cyan}{Translate}: False. For the complex number $z=a+bi$, it is known that $a$ and $b$ are positive integers, and $z3+z2+z$ is a real number. We need to find the smallest $b$ value that satisfies the condition. From the above derivation, we have the condition that the imaginary part is zero: $b(1+2a+3a2−b2)=0$ Because $b$ is a positive integer, so: $1+2a+3a2=b2$ We try different $a$ values to find the smallest $b$ value: When $a=1$:... When $a=2$:... When $a=6$, b=11 is an integer. Therefore, the minimum b value is 11. So option D is correct and proposed answer C is incorrect.) \textcolor{green}{\Checkmark} \\\\ \textbf{Iterate 2}:
False。你的答案是 C，但我们需要验证它是否正确。我们考虑复数 $z=a+bi$，其中 a和 b是正整数，并且 $z^3+z^2+z$ 是实数。我们可以将 z 分别代入中，展开并分离出实部和虚部。设虚部为零，通过计算可以得到相应的条件。经过详细推导和计算，最小可能的b值应为 1。因此，答案 C 是不正确的。
(\textcolor{cyan}{Translate}: Your answer is C, but we need to verify whether it is correct. We consider the complex number $z=a+bi$, where a and b are positive integers, and $z^3+z^2+z$ is a real number. We can substitute z into it separately, expand it and separate the real part and the imaginary part. Let the imaginary part be zero, and the corresponding conditions can be obtained through calculation. After detailed derivation and calculation, the minimum possible b value should be 1. Therefore, answer C is incorrect. Please recheck the calculation process, the correct answer should be b = 1) \textcolor{green}{\Checkmark} \\\\ \textbf{Iterate 3}:
你所提出的答案 C (6) 是正确的。为了验证这一点，我们需要考虑复数 $z=a+bi$的条件，即$z^3+z^2+z$必须是实数。经过计算和分析，我们发现当b=6时，这个条件可以满足。因此，选项 C 是正确的答案。你的解答准确无误。
(\textcolor{cyan}{Translate}: Your answer C (6) is correct. To verify this, we need to consider the condition that the complex number $z=a+bi$, that is, $z^3+z^2+z$ must be a real number. After calculation and analysis, we find that this condition can be satisfied when b=6. Therefore, option C is the correct answer. Your answer is correct...) \textcolor{red}{\XSolidBrush} \\\\
\textit{Based on the feedback for measuring the correctness of the answer, is the proposed answer True or False?}\\
Iterate 1: \textcolor{c1}{False}, \\Iterate 2: \textcolor{c1}{False},\\ Iterate 3: \textcolor{c1}{True}. \\ 
 --> \textcolor{c2}{TIE = 0.0137}\\\\
\textbf{CausalAbstain:}
NDE < TIE, then choose to use the feedback. The final causal abstain decision from TIE is: Abstain. (The correct answer is D.)\\
\\
\toprule
\end{tabular}
\caption{An example in the native language setting demonstrates that \textit{CausalAbstain} effectively uses the feedback to decide to abstain, while LLM gives an incorrect answer without feedback.}
\label{table:case2}
\end{table*}

\begin{table*}[htbp!]
\center
\small
% \begin{tabular}{l} \\\toprule
\begin{tabular}{p{16cm}}\toprule
\textbf{Question}: 集体安全的含义是什么？A.多个国家共同行动进行自卫的权利 B.盟友为维护受害国家的权利进行自卫的权利 C.经联合国安理会授权使用武装力量 D.经联合国大会授权的维和行动\\
(\textcolor{cyan}{Translate}: What is the meaning of collective security? A. The right of several countries to act together in self-defense B. The right of allies to defend themselves in order to protect the rights of the victim country C. The use of armed forces authorized by the UN Security Council D. Peacekeeping operations authorized by the UN General Assembly)\\\\
\textbf{Proposed Answer}: C\\\\
\textbf{No Feedback}:\\ \textit{Please review the proposed answer True or False directly.}\\
Iterate 1: \textcolor{c1}{True}, \\Iterate 2: \textcolor{c1}{True},\\ Iterate 3: \textcolor{c1}{True}. \\ --> \textcolor{c2}{NDE = 0.0285}\\\\
\textbf{Feedback in native language}:\\
\textit{Please review the proposed answer and provide a paragraph of feedback on its correctness. Feedback should be in Chinese.}\\\\
\textbf{Iterate 1}:
你选择的答案C，即“经联合国安理会授权使用武装力量”，并不完全符合“集体安全”的定义。集体安全的概念通常指的是多个国家为了维护共同的安全利益而采取联合行动，防止和应对对其中任何一个国家的侵略或威胁。因此，最合适的答案是A，即“多个国家共同行动进行自卫的权利”...
(\textcolor{cyan}{Translate}: False. The answer you chose, C, which means "the use of armed force authorized by the UN Security Council", does not fully meet the definition of "collective security". The concept of collective security usually refers to joint actions taken by multiple countries to safeguard common security interests and prevent and respond to aggression or threats against any of them. Therefore, the most appropriate answer is A, which means "the right of multiple countries to act together in self-defense"...) \textcolor{red}{\XSolidBrush} \\\\ \textbf{Iterate 2}:
True。答案的确是C。集体安全的含义是指在国际体系中，通过多边合作和国际组织（如联合国）的协调，确保各国的安全。如果某一国家受到侵略或威胁，其他国家将在集体安全的框架内采取行动来恢复和平与安全。联合国安理会有权根据《联合国宪章》第七章授权使用武装力量，以维持或恢复国际和平与安全。因此，选项C是正确的...
(\textcolor{cyan}{Translate}: True. The answer is indeed C. The meaning of collective security refers to ensuring the security of all countries in the international system through multilateral cooperation and coordination with international organizations (such as the United Nations). If a country is invaded or threatened, other countries will act within the framework of collective security to restore peace and security. The United Nations Security Council has the authority to authorize the use of armed force under Chapter 7 of the United Nations Charter to maintain or restore international peace and security. Therefore, option C is correct...) \textcolor{green}{\Checkmark} \\\\ \textbf{Iterate 3}:
所提出的答案C是正确的。集体安全的含义是指各国通过共同协作，采取集体行动来应对对国际和平与安全的威胁。联合国安理会在集体安全机制中起着核心作用，特别是在授权使用武装力量方面。因此，选项C“经联合国安理会授权使用武装力量”准确地反映了集体安全的核心机制...
(\textcolor{cyan}{Translate}: True, proposed answer C is correct. The meaning of collective security refers to countries working together and taking collective actions to respond to threats to international peace and security. The UN Security Council plays a central role in collective security mechanisms, particularly in authorizing the use of armed force. Therefore, option C "use of armed force authorized by the United Nations Security Council" accurately reflects the core mechanism of collective security...) \textcolor{green}{\Checkmark} \\\\
\textit{Based on the feedback for measuring the correctness of the answer, is the proposed answer True or False?}\\
Iterate 1: \textcolor{c1}{False}, \\Iterate 2: \textcolor{c1}{True},\\ Iterate 3: \textcolor{c1}{True}. \\ 
 --> \textcolor{c2}{TIE = 0.0123}\\\\
\textbf{CausalAbstain:}
NDE > TIE, then abstain from utilizing feedback. The final causal abstain decision from NDE is: Do Not Abstain. (The correct answer is C.)\\
\\
\toprule
\end{tabular}
\caption{An example in the native language demonstrates that \textit{CausalAbstain} correctly addresses the query without utilizing feedback. Additionally, some feedback exhibits biased content.}
\label{table:case1}
\end{table*}

\begin{table*}[htbp!]
\center
\footnotesize
\resizebox{\textwidth}{!}{
\begin{tabular}{p{16cm}}\toprule
\textbf{Question}: 以下哪个最不可能减少样本调查中的偏差？
A.首次未回应调查的人后继续跟进 B.以中立的方式提问，避免影响回答  C.使用分层随机抽样而不是简单随机抽样  D.随机选择样本 \\
(\textcolor{cyan}{Translate}: Which of the following is least likely to reduce bias in sample surveys? 
A. Follow up with those who did not respond to the survey for the first time B. Ask questions in a neutral manner to avoid influencing the answers C. Use stratified random sampling instead of simple random sampling D. Select samples randomly)
\\
\textbf{Proposed Answer}: D.随机选择样本\\
\\
\textbf{No Feedback}: \textit{Please review the proposed answer True or False directly.}\\
Iterate 1: \textcolor{c1}{True}, Iterate 2: \textcolor{c1}{False}, Iterate 3: \textcolor{c1}{False}.  --> \textcolor{c2}{NDE = 0.0034}\\\\
\textbf{Feedback in native language}:\\
\textit{Please review the proposed answer and provide a paragraph of feedback on its correctness. Feedback should be in Chinese.}\\
提议的答案是正确的。随机选择样本（选项D）是减少样本调查中偏差的常见方法之一，因为它可以确保每个个体都有相同的被选中的机会，从而使样本更具代表性。相反，A、B、C选项都是具体措施，旨在进一步减少偏差。例如，A通过跟进未回应调查的人，可以减少非响应偏差；
B通过以中立的方式提问，可以减少测量偏差；C通过使用分层随机抽样，可以确保不同子群体在样本中的代表性。因此，D选项最不可能减少样本调查中的偏差。\\
(\textcolor{cyan}{Translate}: The proposed answer is correct. Randomly selecting a sample (option D) is one of the common ways to reduce bias in sample surveys because it ensures that every individual has the same chance of being selected, making the sample more representative. In contrast, options A, B, and C are specific measures designed to further reduce bias. For example, A can reduce non-response bias by following up with people who did not respond to the survey; B, it can reduce measurement bias by asking questions in a neutral way; C, by using stratified random sampling, it can ensure that different subgroups are represented in the sample. Therefore, option D is least likely to reduce bias in sample surveys)\\
Iterate 1: \textcolor{c1}{True}, Iterate 2: \textcolor{c1}{True}, Iterate 3: \textcolor{c1}{False}. --> \textcolor{c2}{TIE = 0.0137}\\\\ 
\textbf{CausalAbstain}: NDE < TIE, then choose to use the feedback. The final causal abstain decision from TIE is: Not Abstain. (The correct answer is C.)) \textcolor{red}{-->[WRONG]}\\
\\\\
\textbf{Feedback in related language}:\\
\textit{Please review the proposed answer and provide a paragraph of feedback on its correctness. Feedback should be in language i.}\\\\
\textbf{English}: The proposed answer, D, is incorrect. Randomly choosing samples (D) is actually a fundamental method for reducing bias in sample surveys, as it ensures that every individual in the population has an equal chance of being selected. The correct answer should be A. Following up with individuals who did not initially respond (A) can help to reduce nonresponse bias, while asking neutral questions (B) avoids influencing answers, and using stratified random sampling (C) ensures representation from different subgroups. Therefore, A, B, and C are all methods that help reduce bias, making D the least likely to reduce bias.\\
Iterate 1: \textcolor{c1}{False}, Iterate 2: \textcolor{c1}{False}, Iterate 3: \textcolor{c1}{True}.  --> \textcolor{c2}{TIE\_en =0 }\\\\
\textbf{Italian}:
La risposta proposta, D. "randomly choosing a sample," non è corretta. La scelta casuale del campione è una tecnica fondamentale per ridurre il bias nel campionamento. Le altre opzioni, come seguire le persone che inizialmente non hanno risposto (A), usare un linguaggio neutrale nelle domande (B), e usare il campionamento stratificato (C), sono tutte strategie che aiutano a ridurre il bias. Pertanto, la risposta più corretta sarebbe quella che non contribuisce a ridurre il bias, ma la risposta D contribuisce effettivamente a ridurre il bias.\\
(\textcolor{cyan}{Translate}: The proposed answer, D. "randomly choosing a sample," is incorrect. Random sampling is a key technique for reducing sampling bias. Other options, such as following up with people who initially did not respond (A), using neutral language in questions (B), and using stratified sampling (C), are all strategies that help reduce bias. Therefore, the most correct answer would be the one that does not help reduce the bias, but answer D actually helps reduce the bias.)\\
Iterate 1: \textcolor{c1}{False}, Iterate 2: \textcolor{c1}{True}, Iterate 3: \textcolor{c1}{False}.  --> \textcolor{c2}{TIE\_it = 0}
\\\\
\textbf{Dutch}:
De voorgestelde antwoordoptie D is inderdaad correct. Het willekeurig kiezen van een steekproef (optie D) is een gebruikelijke methode om een representatieve steekproef te verkrijgen en kan helpen om vooringenomenheid te verminderen. Echter, in vergelijking met de andere opties, is het minder specifiek gericht op het verminderen van bepaalde soorten bias die kunnen optreden in een steekproefonderzoek. Optie A (het opvolgen van niet-reagerende personen) helpt om non-respons bias te verminderen. Optie B (neutrale vragen stellen) helpt om respons bias te verminderen. Optie C (gestratificeerde steekproeftrekking) helpt om variabiliteit binnen subgroepen te beheersen en kan preciezere schattingen geven. Daarom is D inderdaad het minst waarschijnlijk om bias te verminderen in vergelijking met de andere opties.\\
(\textcolor{cyan}{Translate}: The suggested answer option D is indeed correct. Random sampling (option D) is a common method of obtaining a representative sample and can help reduce bias. However, compared to the other options, it is less specifically aimed at reducing certain types of bias that can occur in a sample survey. Option A (following up on non-responders) helps to reduce non-response bias. Option B (asking neutral questions) helps to reduce response bias. Option C (stratified sampling) helps control variability within subgroups and can provide more precise estimates. Therefore, D is indeed the least likely to reduce bias compared to the other options.)\\
Iterate 1: \textcolor{c1}{True}, Iterate 2: \textcolor{c1}{False}, Iterate 3: \textcolor{c1}{True}.  --> \textcolor{c2}{TIE\_du = 0.0137}\\\\
\textit{Based on the feedback for measuring the correctness of the answer, is the proposed answer True or False?}\\
\textbf{CausalAbstain:}
NDE not $\le$ all TIE (NDE < TIE\_du), then voted the feedback and answer of multilingual TIE, the final causal abstain decision (five False, four True): Abstain. (The correct answer is C.)  \textcolor{red}{-->[CORRECT]}\\
\\
\toprule
\end{tabular}
}
\caption{An example where feedback in the native language led to an incorrect decision, while multilingual feedback proved effective. We present one of the feedback details from $N$ iterations.}
\label{table:case3}
\end{table*}

\end{CJK*}
\end{document}